\documentclass[lettersize,journal]{IEEEtai}
\usepackage{amsmath,amsfonts,amssymb}
\usepackage[ruled,vlined,linesnumbered]{algorithm2e}
\usepackage{array}
\usepackage[caption=false]{subfig}
\usepackage{textcomp}
\usepackage{verbatim}
\usepackage{graphicx}
\usepackage{balance}
\usepackage{booktabs}
\usepackage{makecell}
\usepackage{arydshln} 
\usepackage{threeparttable}
\usepackage{url}
\usepackage{color}
\usepackage[colorlinks,urlcolor=blue,linkcolor=blue,citecolor=blue]{hyperref}

\usepackage{amsmath,amssymb,amsfonts}
\usepackage[dvipsnames]{xcolor}
\definecolor{red}{HTML}{e85642}
\definecolor{blue}{HTML}{1280B0}

\begin{document}
\title{Adjusting Logit in Gaussian Form for Long-Tailed Visual Recognition}

\author{Mengke~Li, \IEEEmembership{Member, IEEE},
        Yiu-ming Cheung$^\star$, \IEEEmembership{Fellow, IEEE},
	Yang Lu, \IEEEmembership{Member, IEEE},
	Zhikai Hu, \IEEEmembership{Student Member, IEEE},
	Weichao Lan, \IEEEmembership{Student Member, IEEE},
	Hui Huang, \IEEEmembership{Senior Member, IEEE}
\IEEEcompsocitemizethanks{
\IEEEcompsocthanksitem This work was supported in part by NSFC (62306181, 62376233), Guangdong Basic and Applied Basic Research Foundation (2023B1515120026, 2024A1515010163), DEGP Innovation Team (2022KCXTD025), GRF of RGC (12201321, 12202622, 12201323), RGC SRFS (SRFS2324-2S02), China Fundamental Research Funds for the Central Universities (20720230038), and Xiaomi Young Talents Program. \textit{(Corresponding author: Yiu-ming Cheung)}
\IEEEcompsocthanksitem Mengke Li is with Guangdong Laboratory of Artificial Intelligence and Digital Economy (SZ), Shenzhen, China (e-mail: limengke@gml.ac.cn).
\IEEEcompsocthanksitem Yiu-ming Cheung, Zhikai Hu and Weichao Lan are with the Department of Computer Science, Hong Kong Baptist University, Hong Kong SAR, China (e-mail:\{ymc, cszkhu, cswclan\}@comp.hkbu.edu.hk).
\IEEEcompsocthanksitem Yang~Lu is with the Fujian Key Laboratory of Sensing and Computing for Smart City, School of Informatics, Xiamen University, Xiamen, China (e-mail: luyang@xmu.edu.cn).
\IEEEcompsocthanksitem Hui Huang is with College of Computer Science and Software Engineering, Shenzhen University, Shenzhen, China (e-mail: huihuang@szu.edu.cn).
} }

\markboth{IEEE Transactions on Artificial Intelligence}{M.K Li \MakeLowercase{\textit{et al.}}: Adjusting Logit in Gaussian Form for Long-Tailed Visual Recognition}

\maketitle

\begin{abstract}
It is not uncommon that real-world data are distributed with a long tail.
For such data, the learning of deep neural networks becomes challenging because it is hard to classify tail classes correctly.
In the literature, several existing methods have addressed this problem by reducing classifier bias, provided that the features obtained with long-tailed data are representative enough.
However, we find that training directly on long-tailed data leads to uneven embedding space. That is, the embedding space of head classes severely compresses that of tail classes, which is not conducive to subsequent classifier learning. 
This paper therefore studies the problem of long-tailed visual recognition from the perspective of feature level.
We introduce feature augmentation to balance the embedding distribution. The features of different classes are perturbed with varying amplitudes in Gaussian form.
Based on these perturbed features, two novel logit adjustment methods are proposed to improve model performance at a modest computational overhead.
Subsequently, the distorted embedding spaces of all classes can be calibrated.
In such balanced-distributed embedding spaces, the biased classifier can be eliminated by simply retraining the classifier with class-balanced sampling data.
Extensive experiments conducted on benchmark datasets demonstrate the superior performance of the proposed method over the state-of-the-art ones. 
Source code is available at \href{https://github.com/Keke921/GCLLoss}{\textcolor{blue}{https://github.com/Keke921/GCLLoss}}.
\end{abstract}
\begin{IEEEImpStatement}
Long-tailed visual recognition, a burgeoning field within computer vision, holds profound significance in academic discourse.
It fosters advancements in real-world applications by addressing challenges posed by imbalanced datasets, thereby facilitating improved model generalization.
In this paper, we propose a simple yet effective logit adjustment method, applicable across different models.
Our work provides comprehensive discussions of the proposed method for long-tail learning, considering aspects of optimization and geometric interpretation.
These discussions contribute to a deeper understanding of long-tail learning and a novel approach for enhancing generalization on the test set.
In scholarly pursuits, long-tailed visual recognition underscores the necessity for nuanced and inclusive methodologies, which are pivotal in advancing the frontiers of research in computer vision and artificial intelligence.
\vspace{-6pt}
\end{IEEEImpStatement}
\begin{IEEEkeywords}
Imbalance learning, long-tailed classification, Gaussian clouded logit, logit adjustment
\vspace{-12pt}
\end{IEEEkeywords}

\section{Introduction} \label{sec:intro}
Deep learning methods have achieved better-than-human performance on a variety of visual recognition tasks~\cite{he2016deep, Wang2017NormFace, TianJ23Synergetic} by virtue of the large-scale annotated datasets. 
In general, the success of deep neural networks (DNNs) relies on balanced-distributed data and sufficient training samples. 
That is, the number of samples in each class is basically the same and large enough. 
Unfortunately, from the practical perspective, data collected from the real world would follow a power-law distribution~\cite{kendall1948advanced, ZhangYH23Multi}, which means that a tiny number of head classes occupy large volumes of instances while the vast majority of tail classes each have fairly few samples, showing a ``long tail'' in the data distribution. 
In fact, class importance is independent of the number of training samples. 
In other words, few samples cannot imply the unimportance of the tail classes~\cite{DasW22On}. 
Even more, misclassification of tail classes can have severe consequences, especially in critical applications such as medical diagnosis~\cite{BasuA23Do} or road monitoring~\cite{li2022coda}. 
Therefore, it is important to develop methods that can effectively address the long-tailed distribution of data and improve the recognition performance on tail classes particularly.

In the literature, many researchers have addressed the issue of long-tailed visual recognition by focusing on the classifier level. 
It is well-known that DNN can be decoupled into a feature extractor and a classifier~\cite{DeWang2022FewShot, Defang2022simkd}. 
Recently, Zhou et al.~\cite{bbn20} have conducted empirical studies to demonstrate that the features (also referred to as {\it embeddings\/} interchangeably hereinafter) obtained from the original long-tailed dataset are already sufficiently representative.
Consequently, they shifted their focus to balancing the classifier through two versions of sampling data.
Also, two-stage decoupling methods~\cite{decouple20, Kaidi2019ldam, wang2020devil, DisAli21} have been proposed to obtain a representation in the first stage and then re-train the classifier on balanced sampling data in the second stage. 
These methods obtain the representation by cross-entropy (CE) loss, which, however, leads to a severely uneven distribution of the embedding space, hindering the acquisition of a better classifier. Furthermore, re-training the classifier can only alleviate the classifier bias but cannot adjust the distorted embedding space, which is not conducive to further promoting the model performance.

\begin{figure}[t]
\centering
\hspace{-8pt}
 \includegraphics[width=0.9\linewidth]{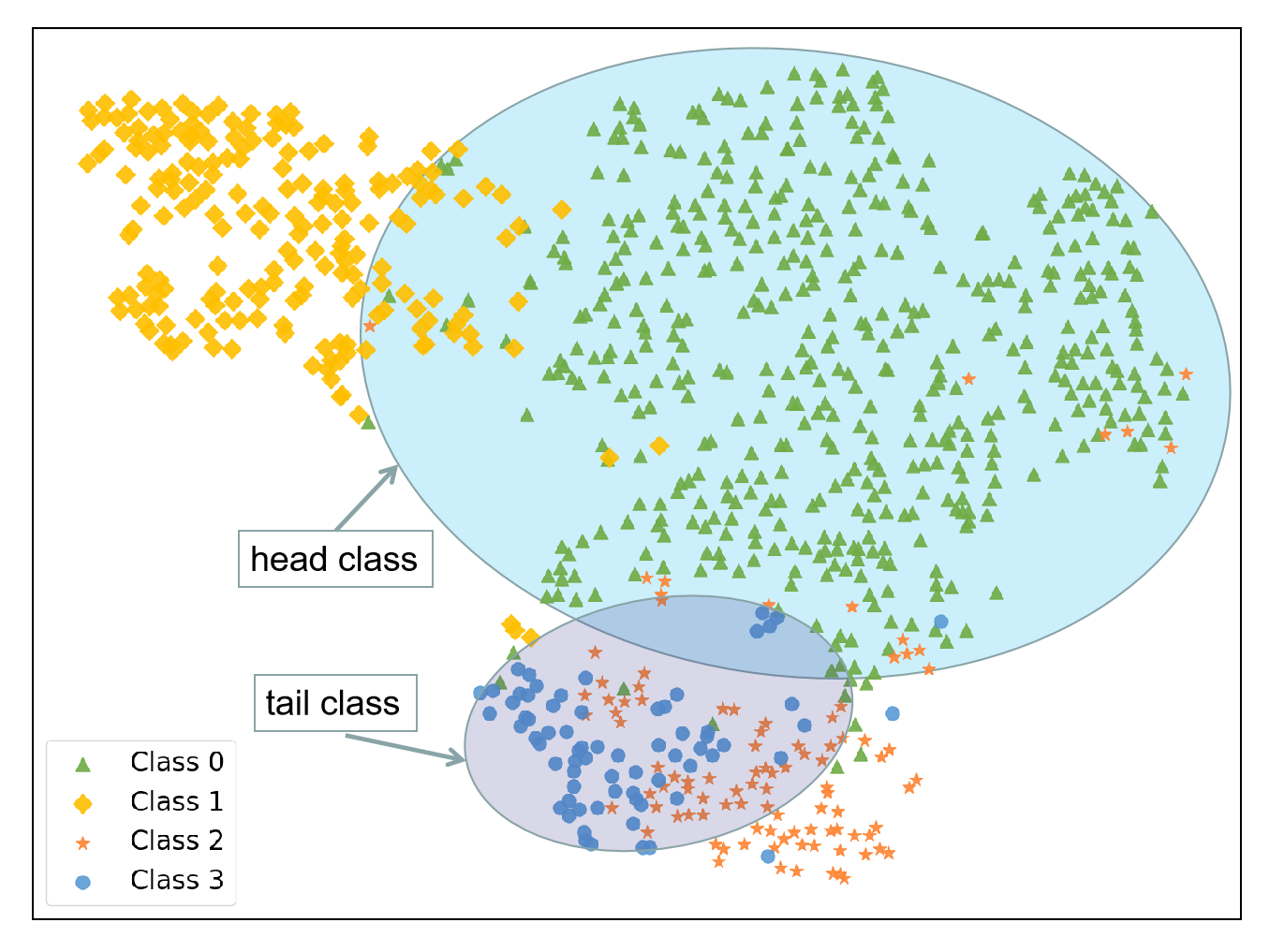}
 \caption{T-SNE visualization of the distorted embedding space\protect\footnotemark. The embedding distributions of head and tail classes are shown in shaded areas. We can see that there are many overlapping regions between each class.}
 \label{fig:intro-toy}
 \vspace{-6pt}
\end{figure}

For the feature issue, specifically, the embedding spatial span of tail classes is drastically compressed by head classes because they have limited training samples that cannot cover the true distribution in embedding space. For ease of understanding, we use a simple experiment to demonstrate the distortion of the embedding space, as illustrated in Fig.~\ref{fig:intro-toy}, where the features are projected by t-SNE~\cite{van2008visualizing}. It can be observed that the tail class occupies a much smaller spatial span than the head class.

\footnotetext{The embeddings are obtained by CE loss from a subset with four classes in CIFAR-10-LT. We randomly select 500, 200, 100, and 50 samples for each class to simulate the data imbalance. }

\begin{figure}[t]
\centering
 \includegraphics[width=0.88\linewidth]{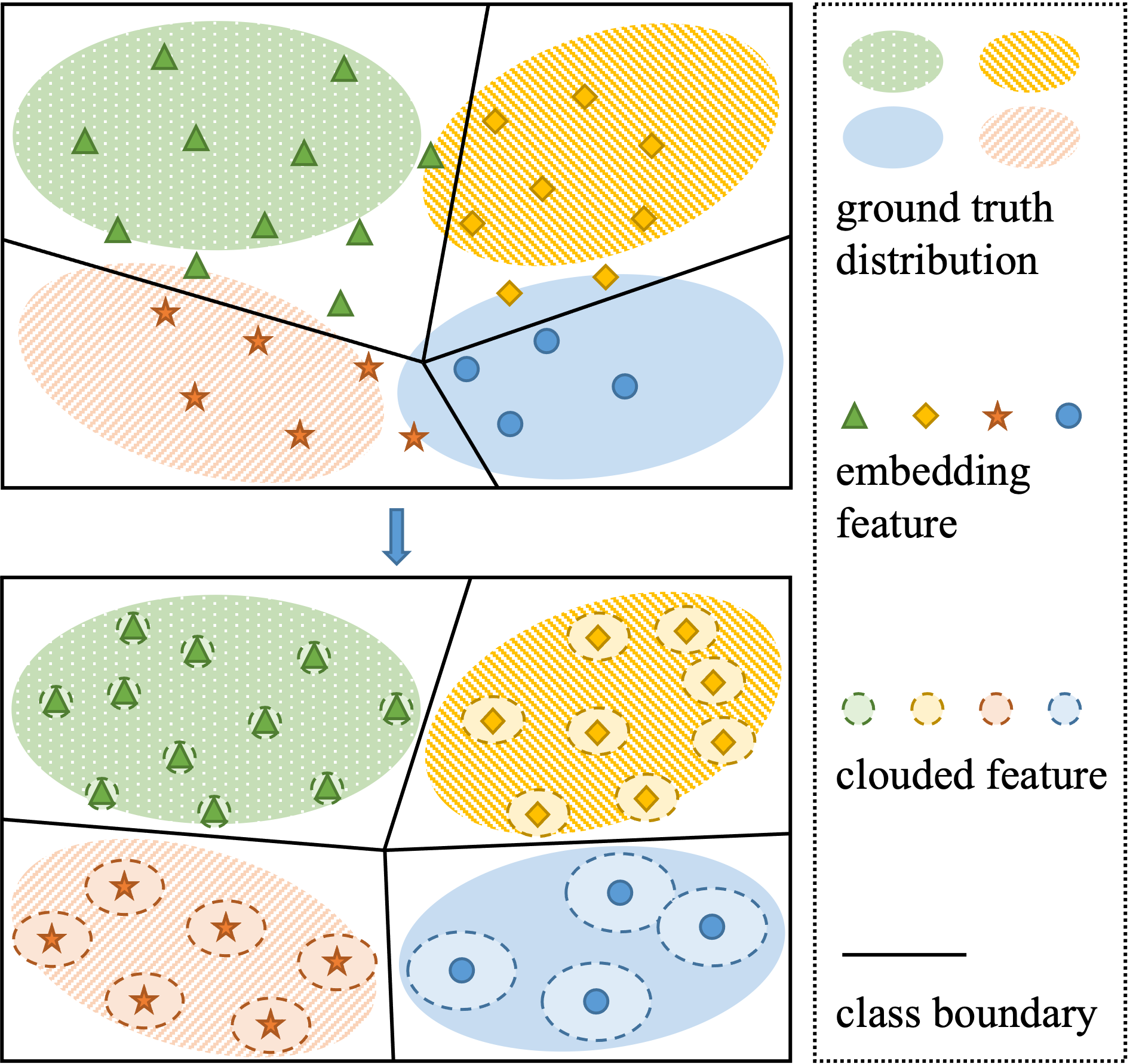}
 \caption{Overview of the proposed method. The embedding distribution obtained by CE loss is uneven, leading to difficulty in classifying the tail class. By assigning larger cloud sizes to the tail class features, the distortion of the embedding space can be well-calibrated.}
 \label{fig:intro}
\end{figure}

A straightforward way to calibrate the distorted embedding space is to enlarge the spatial distribution of tail classes. 
Analogous to human cognition, where a person is capable of inferring the extension of an entire category from a single instance \cite{smith2017developmental}, we treat one training sample as a set of similar samples. 
By augmenting the features, we can control the spatial span of the embedding. 
As only the orientation of the class anchors contributes to the classification, we increase the perturbation amplitude of the tail classes along the direction of the corresponding class anchors.
This expands the spatial distribution of tail classes and prevents them from being overly compressed by head classes.
Conversely, these amplitudes for head classes should be small. Since their samples with enough diversity already cover the actual spacial span, additional expansion is no need anymore. Eventually, as shown in Fig.~\ref{fig:intro}, the tail class samples can be pushed further away from the other classes so that the distortion of the embedding space can be well calibrated. To this end, we first expand the embedding spatial span with a Gaussian form of perturbation. 
Based on this, we propose a novel logit adjustment method in two forms: normalized Euclidean and Angular. This method improves model performance with negligible additional computation. 
Since Gaussian distribution has a cloud-like shape, we name the perturbation amplitude as cloud size and the proposed method as Gaussian clouded logit (GCL). 
After calibrating the embedding space with GCL, the features of different classes can be more evenly distributed. 
It turns out that the classifier bias can be easily eliminated through class-balanced sampling data~\cite{cui2019class, Wang2017Learning} in such a balanced-distributed space. Extensive comparison experiments implemented on multiple commonly used long-tailed benchmarks demonstrate the superiority of the proposed GCL.

Compared to our preliminary work reported in~\cite{CVPR2022}, the primary distinction of this paper can be summarized as follows:
Firstly, this paper provides a general form of perturbed logit by perturbing the logit to calibrate the distribution of embedding space. 
Accordingly, two specific forms based on different metrics are derived from this general form.
Secondly, we present the analysis and explanation of the rationale of GCL in detail, based on which more general parameter selection strategies are provided. 
After calibrating the embedding space with GCL, the classifier bias can be mitigated by simply retraining with the balanced sampling data.
Thirdly, more experiments are conducted to demonstrate the effectiveness of the proposed method. 
Specifically, we add more classification baselines to show the efficacy of GCL. 
Furthermore, we demonstrate that GCL can enhance the performance of mixture of experts (MoE) model.
Additionally, we provide in-depth theoretical and experimental analyses of the characteristics of GCL in both its normalized Euclidean and angular forms.
In summary, the main contributions of this paper are threefold.

\begin{itemize}
    \item[$\bullet$]  We propose a simple but effective GCL adjustment method derived from the Gaussian perturbed feature. Tail classes are assigned larger cloud sizes than head classes along the direction of the corresponding class anchors. Consequently, it can address the problem of the distorted embedding space caused by long-tailed data.
    \item[$\bullet$] We provide in-depth discussions into GCL for long-tail learning from the perspective of optimization and geometric interpretation. They help set the sign and magnitude of the perturbation and provide a new idea for better generalization to the test set.
    \item[$\bullet$] We obtain two specific forms of GCL. Both of them outperform state-of-the-art counterparts on long-tailed benchmark datasets without additional computation. Their advantages and disadvantages in different long-tailed scenarios are analyzed in detail.
\end{itemize}

The remainder of this paper is organized as follows. 
Section~\ref{sec:related_works} makes an overview of the recent related works. 
Section~\ref{method} details the derivation and rational analysis behind the proposed Gaussian clouded logit.
Section~\ref{sec:exp} presents our experimental results in comparison with the baseline methods, as well as model validation and analysis. 
Finally, Section~\ref{sec:conclusion} draws a conclusion.
\section{Related Works}
\label{sec:related_works}
Over the past years, a number of methods have been proposed to address long-tailed visual recognition. This section provides an overview of the most related four regimes. That is, data augmentation, two-stage method, mixture of experts, and loss modification and logit adjustment.
\subsection{Data Augmentation}
Input augmentation increases sample diversity in the data space. 
The classical augmentation methods~\cite{he2016deep} encompass operations such as flipping, rotating, cropping, padding, etc. 
Most recently, Wang~et al.~\cite{Wang2021RSG} proposed rare-class sample generator (RSG) that augments tail classes by utilizing encoded variation information obtained from head classes. 
M2m~\cite{kim2020m2m} establishes a well-balanced dataset through the translation of samples from head classes to tail classes, facilitated by an auxiliary pre-trained classifier.

Feature augmentation serves to enhance data diversity within the feature space.
Knowledge transfer is a promising technology. 
For instance, Yin~et al.~\cite{yin2019feature} exemplified knowledge transfer by leveraging the intra-class variance derived from head classes in an encoder-decoder-based network to augment the features of tail class samples. 
Liu~et al.~\cite{Liu2020Deep} employed the transfer of angular variance, computed from head classes, to enrich the intra-class diversity within tail classes. 
Moreover, recent applications in addressing long-tailed data incorporate the use of class activation maps (CAM)~\cite{Zhou2016Learning}. 
Chu~et al.~\cite{Peng2020Feature} utilized CAM to decompose the features into a class-generic and a class-specific component. Then, tail classes are augmented by fusing the class-specific components obtained from the tail classes with the class-generic components of the head classes. Also, Zhang~et al.~\cite{zhang2021bag} exploited CAM to obtain the foreground in an image and then augment the obtained foreground object by flipping, rotating, jittering, etc. The augmented foreground is then covered on the unchanged background to obtain a new informative image.

Those methods mentioned above require either an increase in data size or model complexity to solve the issues in long-tailed distribution,  resulting in additional computational costs.

\subsection{Two-stage Method}
Recently, two-stage methods have been proposed and empirically demonstrated their efficacy. 
For example, Cao et al.~\cite{Kaidi2019ldam} proposed LDAM-DRW, wherein features are learned in the initial stage, and a deferred re-weighting (DRW) strategy is employed to refine the classifier in the subsequent stage.
While it markedly enhances long-tailed prediction accuracy, the theoretical underpinnings of the deferred DRW strategy remain unclear.
Following this, Kang~et al.~\cite{decouple20} precisely identified out that the learning process of representation and classifier can be decoupled into two separate stages. 
The first stage performs representation learning on the original long-tail data. 
The second stage fixes the parameters of the backbone network and re-trains the classifier using class-balanced sampling data. 
Several studies \cite{wang2020devil, DisAli21, mislas21} have further refined this strategy. 
For example, Zhang et al.~\cite{DisAli21} proposed an adaptive calibration function to calibrate the predicted logits of different classes, aligning them with a balanced class prior to preparation for the second stage.
Zhong et al.~\cite{mislas21} proposed class-based soft labels to address varying degrees of overconfidence in the predicted logit of each class, which can improve the classifier learning in the second stage. 
Another alternative approach is proposed by Zhou et al.~\cite{bbn20}, wherein the network structure is bifurcated into two branches.
One branch focuses on learning the representation of head classes, while the other is tailored for tail classes. 
This structure incorporates feature mixup~\cite{verma2019manifold} into a cumulative learning strategy, yielding state-of-the-art results.
Subsequently, Wang et al.~\cite{contrastive21} introduced contrastive learning into this bilateral-branch structure, further enhancing the performance of long-tailed classification.

\subsection{Mixture of Experts}
More recently, researchers have explored the use of mixture of experts (MoE) methods to enhance performance by integrating multiple models into the learning framework.
The fundamental concept behind these approaches is to introduce diversity to the data or models, which enables experts to concentrate on different portions of the data or allows experts with different structures to analyze the data.
BBN~\cite{bbn20} proposes a two-branched classifier that learns both the long-tailed and inverse distributions simultaneously, with a smooth transition of focus between them.
BAGS~\cite{li2020overcoming}, LFME~\cite{xiang2020learning}, and ACE~\cite{Cai2021ACE} divide the long-tailed data into different sub-splits and fit multiple experts on them.
ResLT~\cite{Cui23ResLT} designs residual structured classifiers that allow experts to specialize in different parts of the long-tailed data and complement each other.
RIDE~\cite{Wang21ride} and TLC~\cite{Li22Trustworthy} employ multiple experts, each trained on different augmented data, to independently learn the long-tailed distribution. 
The predictions of all experts are then gradually integrated to reduce overall model variance or uncertainty.
SHIKE \cite{Jin2023long} investigates the impact of feature depth on data of varying scales in long-tailed visual recognition.
The authors propose a new architecture, which incorporates features from different layers of a neural network to exploit the rich information present at different depths of a network.
NCL~\cite{Li2022CVPR} adopts multiple complete networks to learn the long-tailed data individually and uses self-supervised contrastive strategy~\cite{Cui2021ICCV} to collaboratively transfer knowledge among each individual expert.
\subsection{Loss Modification and Logit Adjustment}
Re-weighting the loss function is one of the most intuitive ways to improve the attention of DNN model on tail classes. 
In the literature, sample-wise re-weighting~\cite{Mengye2018Learning, Tsung2020Focal} introduces the fine-grained coefficients into the loss function to make the model pay more attention to the difficult samples. 
Furthermore, class-wise re-weighting~\cite{Salman2018Cost, cui2019class, tan2020Equalization} assigns the standard CE loss with category-specific parameters that are inversely proportional to the class sizes. 
These methods can alleviate the data imbalance to a certain extent. 
However, when the imbalance ratio is very high, large weights may cause overfitting to the tail classes. 
Besides that, another side effect of assigning higher weights to difficult samples/tail classes is overly focusing on harmful samples (e.g., abnormal samples or mislabeled data)~\cite{Pang2017Understanding}.

Loss function can also be modified by adjusting the logit. 
Menon et al.~\cite{adjustment21} proposed logit adjustment (LA), which is consistent in minimizing the balanced error. The logit shifting in LA of different classes is based on label frequencies of training data. By contrast, LADE~\cite{Hong2021CVPR} post-processes the model prediction by disentangling the training set distribution from the prediction. This method does not require the test set to be a uniform distribution. Also, DisAlign~\cite{DisAli21} adjusts the logit by calibrating the distribution of model prediction to a balanced one by minimizing the expected KL divergence. Overall speaking, these three methods can well adjust the classifier but do not take into account the distorted embedding space. Alternatively, re-margining methods~\cite{Kaidi2019ldam, Cao2020CVPR, Li2022pamikey} address long-tailed data by leaving large relative margins for tail classes during training. For example, label-distribution-aware margin (LDAM) loss~\cite{Kaidi2019ldam} utilizes Rademacher complexity to theoretically prove that the margin should be inversely proportional to a quarter power of class sizes. The hard margin on target logit helps make the samples within a class more compact but the strict margin constraints increase the risk of overfitting and cannot actually expand the tail class coverage area in embedding space.
\section{Proposed Method}\label{method}
The basic idea of our proposed method is to perturb the features with varying magnitudes in the directions of different class anchors, thereby automatically balancing the spatial span of head and tail classes. The details of the proposed approach are presented as follows.

\subsection{Basic Notations}\label{sec:notations}
This section defines the notation used throughout this paper.

\noindent\textbf{For dataset:}
Suppose $\{x,y\}\in \mathcal{T}$ represents a sample $\{x,y\}$ from the training set $\mathcal{T}$, where $\mathcal{T}$ has $C$ classes and $N$ training samples in total, $x$ represents the image that needs to be classified and $y \in \{1,\ldots, C\}$ is the ground truth label. The number of training samples of class $j, (j=\{1,2,\cdots, C\})$ is $n_j$ and $\sum_{j=1}^C{n_j} = N$.

\noindent\textbf{For backbone:}
The feature vector $\mathbf{f} \in \mathbb{R}^D$ is derived from the embedding layer, with a dimensionality of $D$.
$\mathbf{W}=\{\mathbf{w}_1, \mathbf{w}_2,\cdots,\mathbf{w}_C\} \in \mathbb{R}^{D\times C}$ represents the weight matrix of the classifier, where $\mathbf{w}_j$ represents the anchor vector of class $j$ in the classifier. The predicted logit of class $j$ is represented by $z_j$, thus, $z_j = \mathbf{w}_{j}^T \mathbf{f}$. The subscript $y$ indicates the target class. That is, $z_{y}$ denotes the target logit and $z_{j}, j\neq y$ is the non-target logit.

\subsection{Embedding Space Calibration}\label{sec:sc}
Suppose a feature point and a small area around it belong to the same type. It is reasonable that the adjacent points around a feature can be regarded as similar to it, and can naturally be considered as the same class.
\subsubsection{General form via perturbing the embedding representation}
We sample a set of points by adding perturbations following a specific distribution to a given feature. Then, a perturbed feature $\mathbf{f}^{ptb}$ of the input is represented as:
\begin{equation}
    \mathbf{f}^{ptb} \triangleq  \mathbf{f}+\delta\mathbf{E},
\end{equation}
where $\mathbf{E}$ represents the perturbation and $\delta>0$ is the amplitude of it. To avoid misleading the final classification, the perturbation amplitude cannot be too large, thus $\delta$ should be a small number. 
This perturbed feature is the input of the classifier. Then, the corresponding perturbed logit $z^{ptb}_j$ of class $j$ is calculated by:
\begin{equation}\label{eq:disturbed_z_1}
\begin{array}{lll}
  z^{ptb}_j &= \mathbf{w}^T_{j}\mathbf{f}^{ptb} + \mathbf{b}_{j}\\
  &= \mathbf{w}^T_{j}\mathbf{f} + \mathbf{b}_{j} + \mathbf{w}^T_{j}(\mathbf{\delta\mathbf{E}})\\
  &= z_{j} + \delta(\mathbf{w}^T_{j} \mathbf{E}), \\
\end{array}
\end{equation}
where $z^{ptb}_j$ is the original logit $z_{j}$ augmented by a perturbing a perturbing item $\delta(\mathbf{w}^T_{j} \mathbf{E})$.

\subsubsection{Normalized Euclidean form}
It should be noted that the perturbing item has different degrees of influence on the final predicted results based on different predicted logits. 
The impact on $z^{ptb}_j$ is relatively minor when the original logit $z_j$ is large.  
Conversely, it becomes more pronounced for $z^{ptb}_j$ when $z_j$ is small. 
Consequently, it is imperative to normalize the effects induced by varying predicted logits while preserving the consistency of the perturbing item's influence. 
We achieve this by employing cosine distance through the normalization of the perturbed logits. 
Here, $s_e$ and $s_a$ represent the norms of the embedding and the class anchor, respectively, that is $s_e = \|\mathbf{f}\|$ and $s_a = \|\mathbf{w}_j\|$.
The normalized perturbed logit $\Tilde{z}^{ptb}_j$ is expressed as:
\begin{equation}\label{eq:cld_nor_scr}
\begin{array}{ll}
  \Tilde{z}^{ptb}_{j}&=\dfrac{s_a\mathbf{w}^T_{j}\cdot s_e\mathbf{f}^{ptb}}{\|\mathbf{w}^T_{j}\|\|\mathbf{f}^{ptb}\|} \\
  & = \tilde{s} \cdot(\dfrac{\mathbf{w}^T_{j}\mathbf{f}}{\|\mathbf{w}^T_{j}\|\| \mathbf{f}+\delta\mathbf{E}\|} + \delta\dfrac{\mathbf{w}^T_{j}\mathbf{E}}{\|\mathbf{w}^T_{j}\|\| \mathbf{f}+\delta\mathbf{E}\|})
\end{array},
\end{equation}
where $\tilde{s} = s_a \cdot s_e$.
$\|\mathbf{f}+\delta\mathbf{E}\|$ is approximate to $\| \mathbf{f} \|$ because $\delta$ is a small number. 
For the second term, we use $\mathbf{I}_j$ to represent the identity vector that has the same direction as $\mathbf{w}^T_{j}$, namely $\mathbf{I}_j=\dfrac{\textbf{w}^T_j}{\|\textbf{w}^T_j\|}$. 
Eq.~(\ref{eq:cld_nor_scr}) is simplified as:
\begin{equation}
\begin{array}{ll}\label{eq:z_j2}
  \tilde{z}^{ptb}_{j} &\approx \tilde{s} \cdot(\dfrac{\mathbf{w}^T_{j}\mathbf{f}}{\|\mathbf{w}^T_{j}\|\| \mathbf{f}\|} + \delta \mathbf{I}_j \dfrac{\mathbf{E}}{ s_e }) \\
  &= \tilde{s}\cdot(\cos \theta_j+ \dfrac{\delta}{ s_e } \mathbf{I}_j\mathbf{E})
\end{array},
\end{equation}
where $\theta_j$ is the angle between $\mathbf{f}$ and $\mathbf{w}_j$. 
Inspired by \cite{Deng2019ArcFace}, the predictions can be made solely based on the angle between the feature and the class anchor. 
Therefore, following \cite{Wang2017NormFace, wang2018cosface}, we can utilize a fixed norm of individual class anchor to substitute $s_a$. 
Without loss of generality, we employ $s_a = 1$.  
Additionally, following \cite{Ranjan2017L2constrainedSL, wang2018additive, Deng2019ArcFace}, the norm of the embedding feature can also be replaced with a constant $s$, that is, set $s_e = s$.
Consequently, the logit is calculated using features distributed on a hypersphere of radius $s$.
As for the perturbation, we set it to Gaussian distribution, i.e. $\mathbf{E}\sim\mathcal{N}(\mathbf{M},\Sigma)$ where $\mathbf{M} \in \mathbb{R}^D$ and $\Sigma \in \mathbb{R}^{D\times D}$. 
The rationale behind this choice lies in the widespread adoption of additive Gaussian noise in machine learning~\cite{Kim2019Make} attributed to the simplicity and universality~\cite{Rasmussen1999infiniteGMM, Mendenhall2012prosta} of Gaussian distribution.
Moreover, we specifically set $\Sigma=\sigma \mathbf{I}$ where $\mathbf{I} \in \mathbb{R}^{D \times D}$ is the identity matrix. 
Then $\mathbf{I}_j\mathbf{E}$ is the projection of the perturbation on the direction of the anchor vector of class $j$. 
We directly use $\varepsilon_j$ to represent this value, which can be interpreted as the amplitude of the projection.
By substituting the aforementioned norms and perturbation into Eq.~(\ref{eq:z_j2}) and uniformly shifting the class-related variable to the pre-defined perturbation amplitude $\delta$ for simplicity, we derive a more concise expression for $\Tilde{z}^{ptb}_{j}$:
\begin{equation}\label{eq:z_j3}
\begin{array}{cc}
\Tilde{z}^{ptb}_{j}&= s\cdot(\cos \theta_j + \dfrac{\delta}{s}\varepsilon_j)   \\
\\[-9pt]
&\Leftrightarrow s\cdot(\cos \theta_j + \delta_j\varepsilon).
\end{array}
\end{equation}
Since $\varepsilon$ is also distributed in Gaussian form, it has a cloud-like shape. $\delta_j$ is the class-based perturbation amplitude that depends on label frequencies. 
We name $\delta_j$ cloud size because it controls the amplitude of $\varepsilon$. 
To broaden the embedding space for the tail classes, the cloud size for tail classes is required to be larger than that of the head classes. 
Therefore, $\delta_j$ is negatively correlated with $n_j$.
In addition, given that $\cos \theta_j \in \left[-1,1 \right]$, the consistency of the influence of the perturbing item can be maintained.

As $\varepsilon$ makes the logit has a cloud-like shape, we name the perturbed logit as \textit{Gaussian clouded logit} (GCL). 
We delve into Eq.~(\ref{eq:z_j3}). 
If $\varepsilon>0$, $\Tilde{z}^{ptb}_{j}$ corresponds to the points that are closer to the anchor vector of class $j$. 
The correct classification of proximal points does not guarantee the accurate classification of distant points within the same class. 
Therefore, $\varepsilon>0$ will not be helpful for classification. 
On the contrary, a reduced logit corresponds to the points that are relatively far from the class anchor. If the relatively distant points can be predicted correctly, the closer one will definitely be able to assign the right label. The points in the same class that are relatively far from the class anchor should be focused on. $\varepsilon$ therefore should always be negative. We name this logit as GCL in normalized Euclidean form (GCL-E for short) because it is derived from normalized Euclidean distance metric. We modify the perturbed logit and use $\Tilde{z}^{GCL-E}_j$ to represent it, which is expressed as:
\begin{equation}\label{eq:gcl_ned}
\Tilde{z}^{GCL-E}_j=  s\cdot\left(\cos \theta_j  - \delta^{E}_j\|\varepsilon\|\right),
\end{equation}
where $\delta^{E}_j$ is the cloud size for GCL-E.

\subsubsection{Angular form} \label{sec:dev_logit}
The final logit of GCL in normalized Euclidean form is equivalent to adding a class-based perturbation on cosine logit. 
From another perspective, namely metric learning, Eq.~(\ref{eq:gcl_ned}) corresponds to adding a Gaussian form margin with class-based variance to the cosine logit (Section~\ref{sec:metric} provides a detailed analysis). 
Inspired by Deng~et al.~\cite{Deng2019ArcFace}, this Gaussian form margin can also be introduced into the angular distance metric. 
For the sake of distinguishing from GCL-E, this version of GCL is named GCL in Angular form (GCL-A for short). 
Using $\Tilde{z}^{GCL-A}_j$ to represent.
These two forms can be unified into a single expression:
\begin{equation}\label{eq:gcl_uni}
    z^\star_j = s \cdot \left[  \cos \left( \theta_j + \nu^{A} \delta_j^A\|\varepsilon\| \right) - \nu^{E}\delta_j^E\|\varepsilon\| \right],
\end{equation}
where $\nu^{A} \in \{0,1\}$ and $\nu^{E} \in \{0,1\}$ are the switch parameters. 
\begin{itemize}
    \item When $\nu^{A}=1$ and $\nu^{E}=0$, we obtain the Angular form, expressed as follows: \begin{equation}\label{eq:gcl_and}
    \Tilde{z}^{GCL-A}_j=  s\cdot \cos(\theta_j + \delta^{A}_j\|\varepsilon\|),
    \end{equation}
    \item When $\nu^{A}=0$ and $\nu^{E}=1$, we obtain the normalized Euclidean form, denoted as $\Tilde{z}^{GCL-E}_j$, as expressed in Eq.~(\ref{eq:gcl_ned}).
\end{itemize}

By taking the Gaussian clouded logit into the original softmax, we obtain the final loss function of GCL:
\begin{equation}\label{eq:gcl}
  \mathcal{L}^*_{GCL} = -\dfrac{1}{N}\sum_i\log \frac{e^{\Tilde{z}^{GCL}_{y_i}}}{\sum_j e^{\Tilde{z}^{GCL}_j}},
\end{equation}
where $\Tilde{z}^{GCL}_j$ can be the logit of GCL-E ($\Tilde{z}^{GCL-E}_j$) or GCL-A ($\Tilde{z}^{GCL-A}_j$). $\mathcal{L}_{GCL-E}$ is utilized to represent the loss function of GCL-E and $\mathcal{L}_{GCL-A}$ denotes that of GCL-A.

\subsection{Classifier Re-balance}\label{sec:re-balance-cls}
Although both GCL-E and GCL-A calibrate the distorted embedding space well, the problem of classifier bias still remains to be addressed.

In the following, we analyze the reasons for the biased classifier. 
Eq.~(\ref{eq:partial_sm}) implies that the sample of the target class $y$ punishes the classifier weights $\textbf{w}_j$ of non-target class $j, j\neq y$ w.r.t. $p_j$. 
In general, the number of training instances in head classes is enormously greater than in tail classes. 
Therefore, the classifier weights of tail classes receive much more penalty than positive signals during training. 
Consequently, the classifier will be biased towards the head classes and the predicted logits of the tail classes will be seriously suppressed, resulting in low classification accuracy of the tail classes \cite{tan2020Equalization, Wang21Seesaw, Wang21Adaptive}. 
We call this problem of the cross-entropy loss function in long-tailed learning \textit{negative gradient over-suppression}. 
A straightforward approach to cope with it is to make the sample numbers of each class equal~\cite{OchalM23Few} to balance the negative gradients. 
To achieve this goal, we can make the tail classes over-sampling and then re-train the classifier. 
The sampling rate of each class is $ \frac{1}{C}$. Then, the class-balanced sampling rate $q^{cb}_j$ of each sample $x$ from class $j$ is calculated by:
\begin{equation}\label{eq:sam_cb}
    q^{cb}_j = \dfrac{1}{C \cdot n_j}.
\end{equation}
This strategy is called classifier re-training (cRT)~\cite{decouple20}. 
It can also be combined with the \textit{effective number}~\cite{cui2019class}. 
We can replace the actual sample number $n_j$ of class $j$ with the so-called \textit{effective number} $n^{en}_j$, the effective sampling rate $q^{en}_j$ of each sample from class $j$ is given by:
\begin{equation}\label{eq:sam_en}
    q^{en}_j = \dfrac{1}{C \cdot n^{en}_j},
\end{equation}
where $n^{en}_j$ is calculated by:
\begin{equation}\label{eq:en_sr}
    n^{en}_j = \dfrac{1-\beta^{n_j}}{1-\beta} N,
\end{equation}
with hyper-parameter $\beta \in [0,1)$. Algorithm~\ref{al:gcl} summarizes the overall training procedure of the proposed method.

\begin{algorithm}[t]
\caption{GCL with cRT}\label{al:gcl}
\SetAlgoLined
\KwIn {Training dataset $\mathcal{T}$\;}
\KwOut {Predicted labels\;}
Initialize the model parameters $\omega$ of the backbone network $\phi((x,y);\omega)$\ randomly \;
\For {$iteration=1$ to $I_0$}{
Sample a batch of data $\mathcal{B}$ from the original long-tailed dataset $\mathcal{T}$ with a batch size of $b$\;
Calculate the logit cloud size $\delta_j$ by Eq.~(\ref{eq:delta_exp}) (or Eq.~(\ref{eq:delta_log})):
\qquad\qquad\qquad $\delta_j \leftarrow n_{max} \cdot n_j^{-k} $ (or $\delta_j \leftarrow \log n_{max}-\log n_j$)\;
Calculate the loss by Eq.~(\ref{eq:gcl}):
\qquad$\mathcal{L}((x,y);\omega) = \frac{1}{b}\sum_{(x,y)\in \mathcal{B}} \mathcal{L}^*_{GCL}(x,y)$\;
Update model parameters:
\qquad$\omega = \omega - \alpha \nabla_{\omega} \mathcal{L}((x,y);\omega)$.
}
\For{$iteration=I_0 + 1 $ to $I_0+I_1$}{
Calculate sampling rate by Eq.~(\ref{eq:sam_cb}) (or Eq.~(\ref{eq:sam_en})):
\qquad\qquad\qquad\qquad\qquad $q_j \leftarrow {n_j}/{\sum{n_j}}$ (or $q_j \leftarrow {n^{en}_j}/{\sum{n^{en}_j}} $)\;
Sample a batch of data $\mathcal{B'}$ with the sampling rate $q_j$ and the batch size $b$\;
Calculate the loss using Eq.~(\ref{eq:gcl}):
$\mathcal{L}((x,y);\omega) = \frac{1}{b}\sum_{(x,y)\in \mathcal{B'}} \mathcal{L}_{GCL}(x,y)$\;
Update the classifier parameters $\omega_{cls}$ while keeping the representation parameters frozen:
$\omega_{cls} = \omega_{cls} - \alpha \nabla_{\omega_{cls}} \mathcal{L}((x,y);\omega_{cls})$.
}
\end{algorithm}

\section{Rationale Analysis}\label{sec:rationale}
This section provides a detailed rationale analysis of how Eq.~(\ref{eq:gcl_uni}) and Eq.~(\ref{eq:gcl_and}) balance the embedding space from two perspectives, considering both model optimization and metric learning perspectives, following with a time-complexity analysis.

\subsection{The Perspective of Model Optimization}
In backward propagation, the gradients on logit $z_i$ are calculated by:
\begin{equation}\label{eq:partial_sm}
\frac{\partial \mathcal{L}}{\partial z_i} =
\left\{
\begin{array}{lr}
p_i-1, &i = y\\
p_i,   &i\neq y
\end{array},
\right.
\end{equation}
where $p_i = \dfrac{e^{z_{i}}}{\sum_{j=1}^C e^{z_j}}$. We take the binary case to illustrate without loss of generality. Suppose the input image is from class 1. The gradient on $z_1$ is calculated by:
\begin{equation}\label{eq:bi_partial}
\frac{\partial \mathcal{L}}{\partial z_1} = -\frac{1}{1+e^{z_1-z_2}}.
\end{equation}
It indicates that the gradient of the target class rapidly approaches zero with the increase of the target logit. This phenomenon is called softmax saturation~\cite{Chen2017CVPR, Zhang2021class}. This inopportune early gradient vanishing weakens the validity of training samples and impedes model training. Therefore, softmax can only slightly separate various classes, and lacks the impetus to evenly distribute each class in the embedded space. We can also observe that there are many overlapping areas among each class in Fig.~\ref{fig:intro-toy}. Especially under the circumstances of long-tailed classification, the tail class features are insufficient to cover the ground truth distribution in embedding space. The early gradient vanish caused by soft saturation exacerbates the squeezing of the embedding distribution in tail class.

Different from the original softmax loss function, the logit difference ($\Delta_{y-j}$) obtained by GCL of Eq.~(\ref{eq:gcl_ned}) between the target and non-target classes is calculated by:
\begin{equation}
\begin{array}{ll}
\Delta_{y-j} & =\tilde{z}^{GCL-E}_y-\tilde{z}^{GCL-E}_j\\
& = s \cdot\left(\cos \theta_y-\cos \theta_j-({\delta^E_y-\delta^E_j}) \|\varepsilon\|\right).
\end{array}
\end{equation}
In case the target class is a tail class, $\delta_y-\delta_j>0$, which decreases the softmax saturation and thereby helps increase the validity of tail class samples. Eq.~(\ref{eq:gcl_and}) has the same effect. Thus, Eq.~(\ref{eq:gcl_ned}) and Eq.~(\ref{eq:gcl_and}) can automatically balance the sample validity of different classes and provide incentives for the model to make each class more separable. They achieve the aim of calibrating the distorted embedding space.

\subsection{The Perspective of Metric Learning} \label{sec:metric}
\begin{figure}
  \centering
  \subfloat[Hard margin]{
  \hspace{-6pt}
        \includegraphics[width=0.46\linewidth]{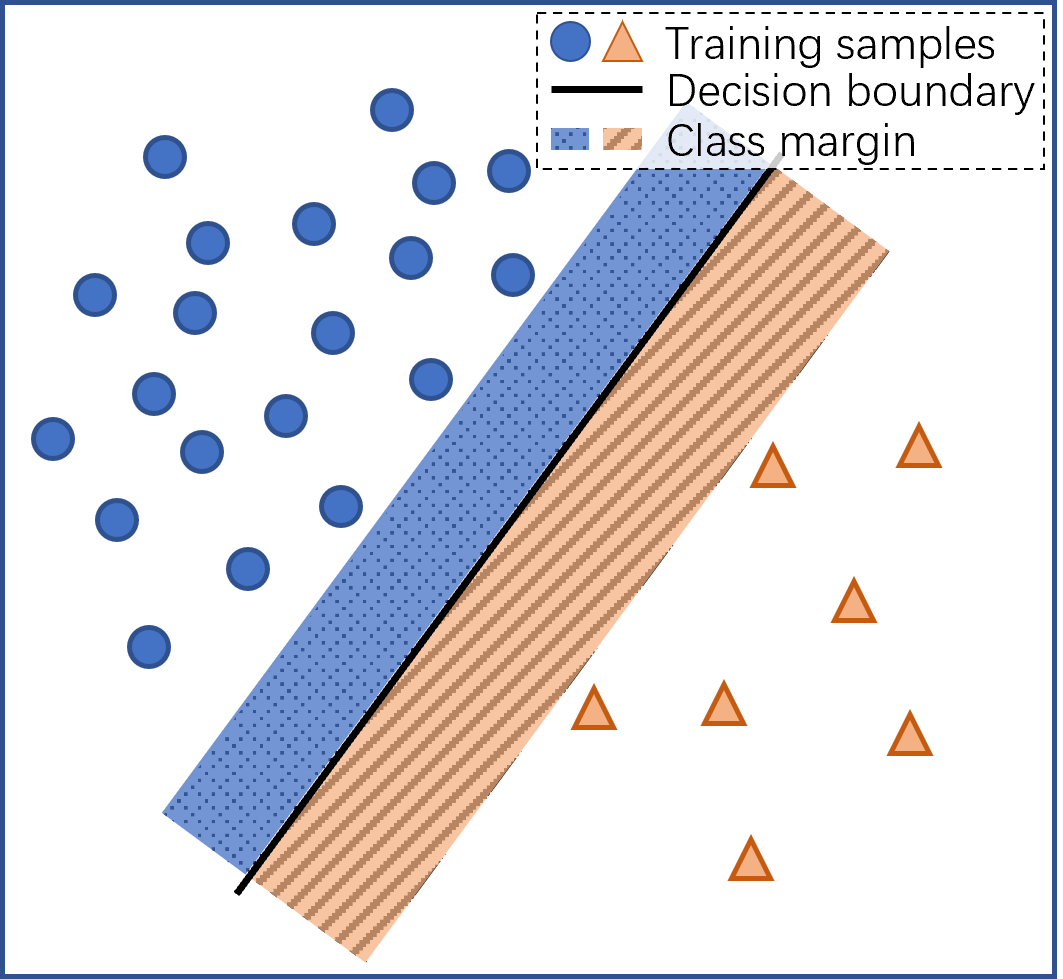}
        \label{fig:pri_mar}}
  \subfloat[Soft margin]{
        \includegraphics[width=0.46\linewidth]{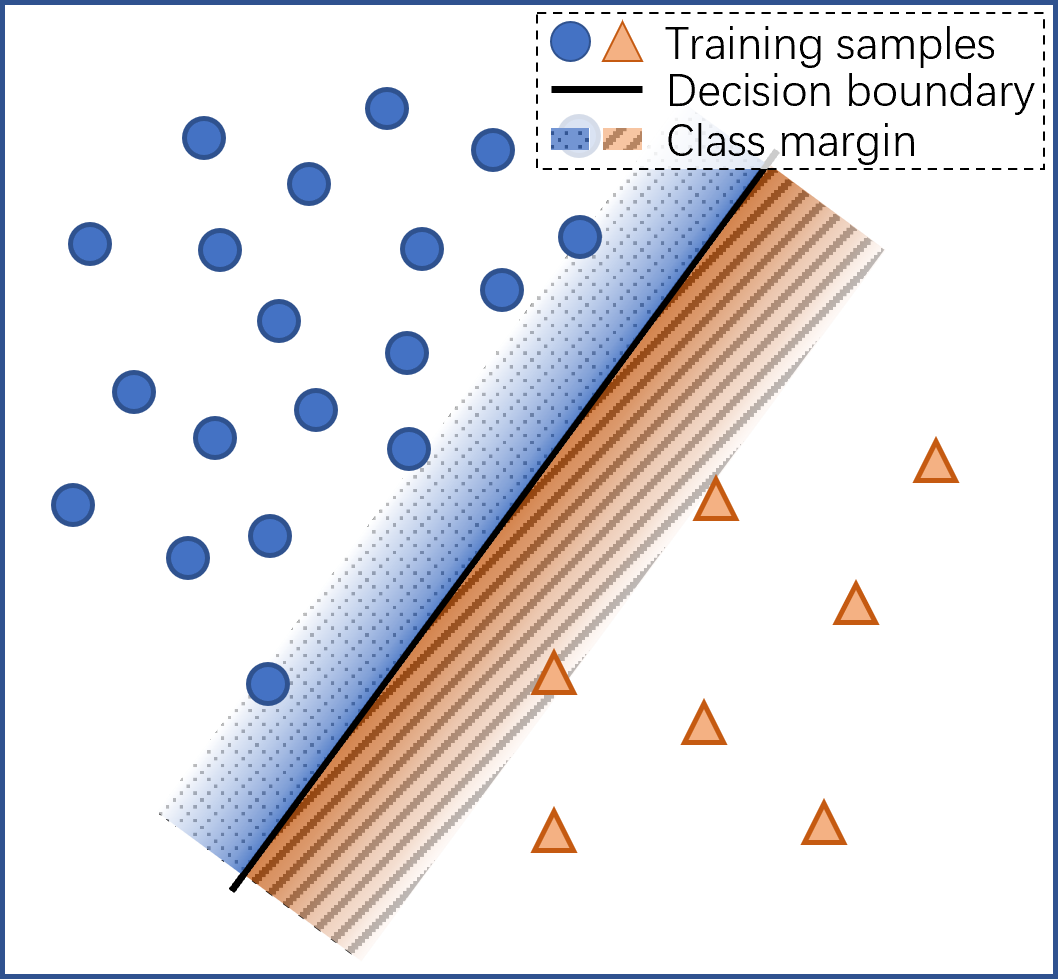}
        \label{fig:gcl_mar}}
\caption{Schematic comparison of hard margin and soft margin. The blue dots and pink triangles represent the head and tail classes, respectively. (Color for the best view.) (a) The hard margin strictly restricts samples from appearing in the corresponding region. (b) The soft margin allows outliers to appear in the region with a lower probability, which increases generalization. }
\label{fig:com_mar}
\end{figure}

Compared with the prior work that enlarges the inter-class separability via the ``hard margin'', e.g. see~\cite{Deng2019ArcFace, Kaidi2019ldam, Zhang2021class}, Eq.~(\ref{eq:gcl_ned}) and Eq.~(\ref{eq:gcl_and}) are equivalent to adding a ``soft'' margin. That is, the farther away from the class anchor, the lower the probability that the point belongs to this class. Fig.~\ref{fig:com_mar} schematically shows the comparison of the prior hard margin and the proposed soft margin. Hard margins will cause the samples to shrink toward the class anchor if the margin is too large. In addition to this, hard margins can lead to overfitting because they prohibit outliers, which can impair the robustness ability of the model. 
The proposed soft margin provides a smooth transition area, allowing the outliers to appear near the target class with a lower probability. 
This is both intuitively and theoretically more reasonable.  

The cloud size $\delta^{*}_j$ may also take different expression forms, where the superscript $*$ indicates the adopted specific form. Cao et al.~\cite{Kaidi2019ldam} obtained the optimal trade-off of the hard margin ($m_i$) and the class size via Rademacher complexity. 
They have proved that $m_i\propto n_i^{-1/4}$. 
The exponent should be ${-1/3}$ derived from Wei et al.~\cite{Wei20Improved}. 
Inspired by these works, we can set the cloud size in power function form:
\begin{equation}\label{eq:delta_exp}
\delta^{pow}_j = n_{max} \cdot n_j^{-k},
\end{equation}
where $n_{max}$ is the sample number of the most frequent class. $k$ can be ${1/3}$ or ${1/4}$. Menon~et al.~\cite{adjustment21} used the Fisher consistency with respect to the balanced error and obtained that $m_i\propto \log (1/n_j)$. Therefore, we can also set the cloud size in logarithmic form:
\begin{equation}\label{eq:delta_log}
\delta^{log}_j = \log n_{max}-\log n_j.
\end{equation}
We also experimentally demonstrate the effectiveness of the cloud size in different expression forms in Section~\ref{sec:cloud_size}.

\begin{figure}
  \centering
  \subfloat[CE]{
  \hspace{-6pt}
        \includegraphics[width=0.23\linewidth]{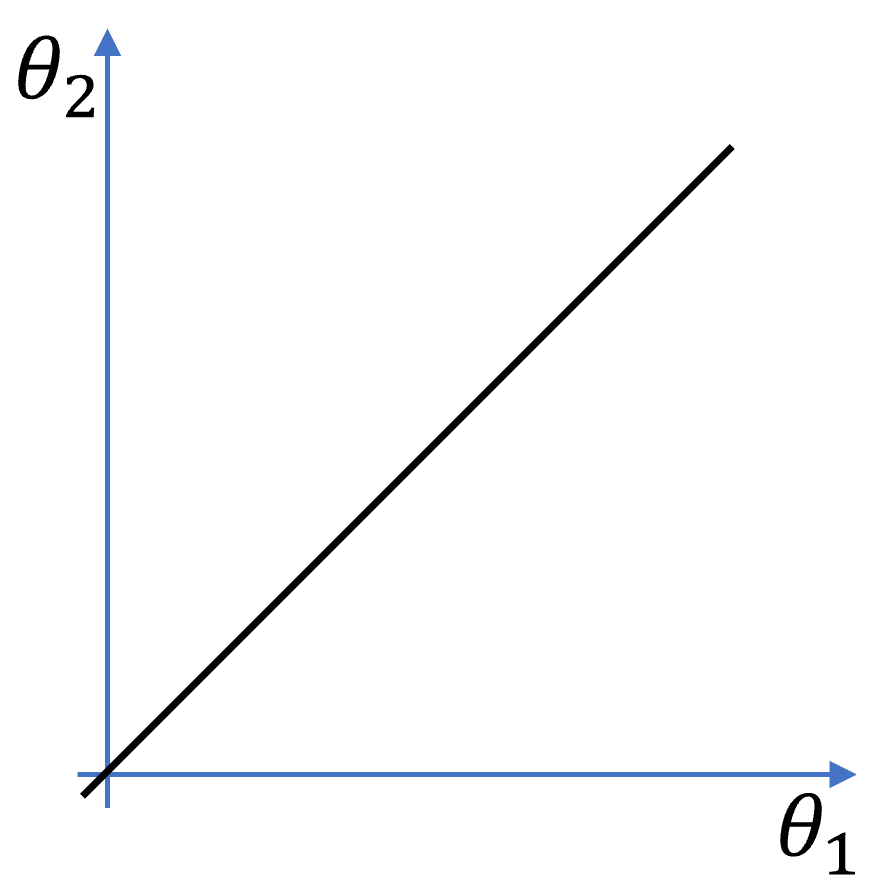}
        \label{fig:CE_mar}}
  \subfloat[LDAM]{
        \includegraphics[width=0.23\linewidth]{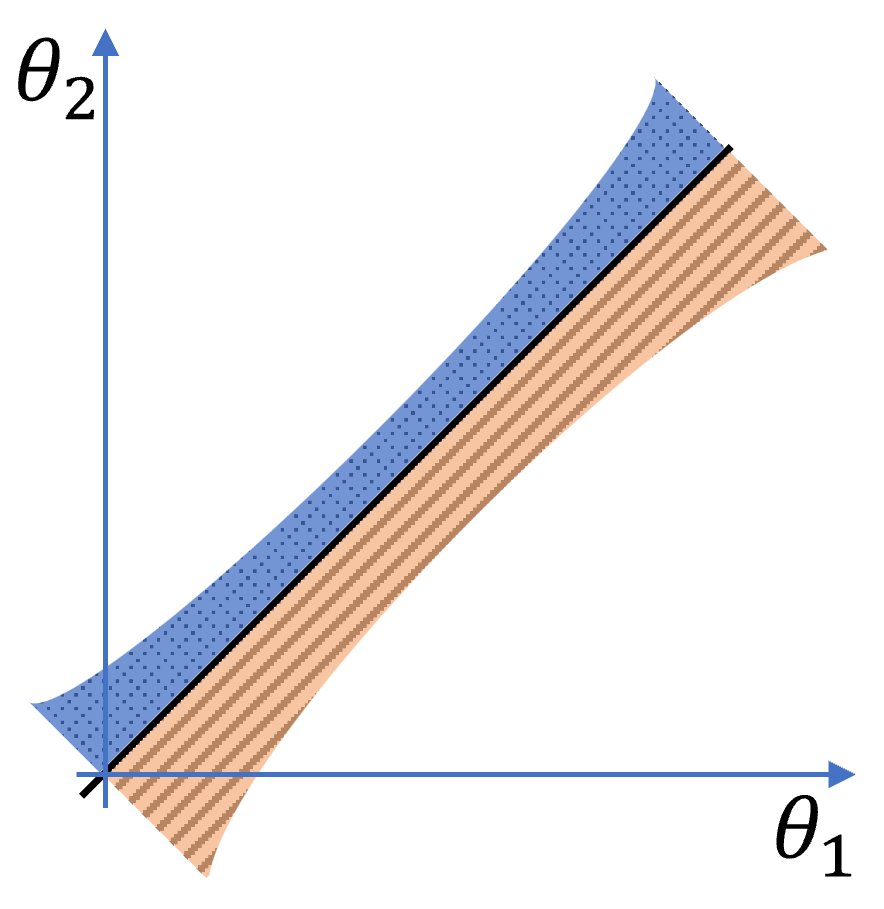}
        \label{fig:LDAM_mar}}
  \subfloat[GCL-E]{
        \includegraphics[width=0.23\linewidth]{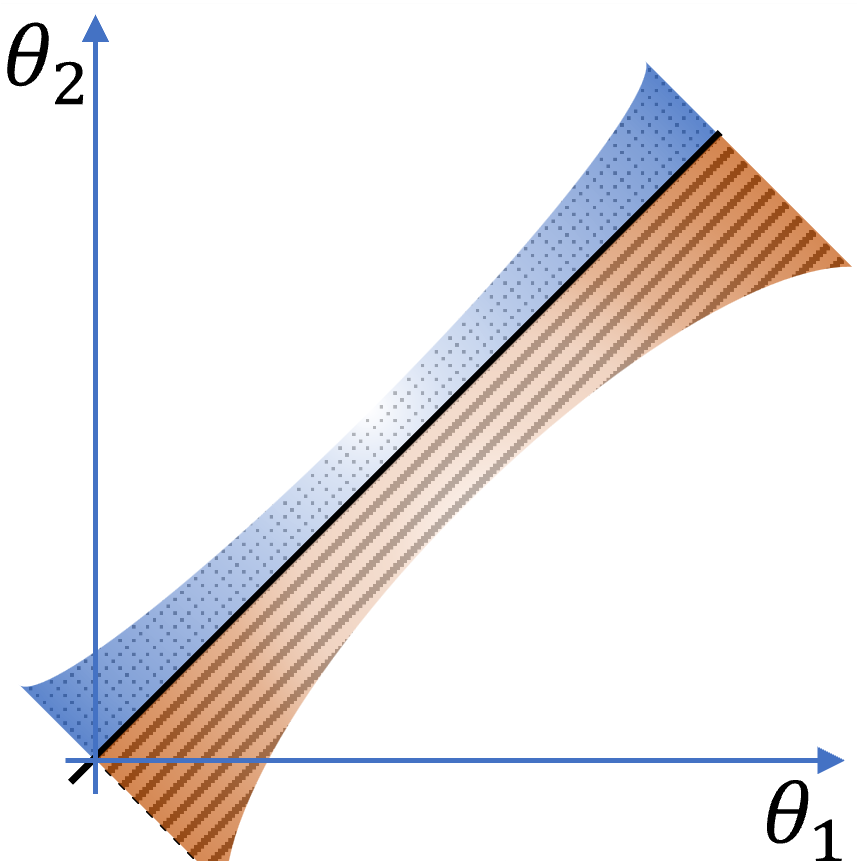}
        \label{fig:GCL-E_mar}}
  \subfloat[GCL-A]{
        \includegraphics[width=0.23\linewidth]{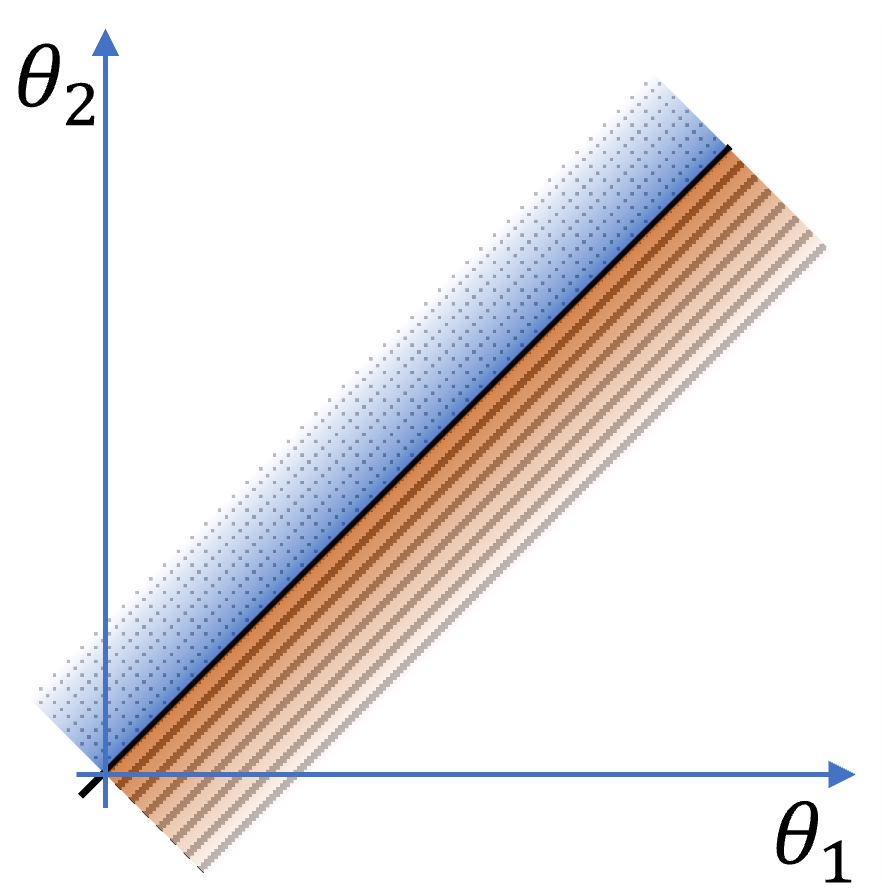}
        \label{fig:GCL-A_mar}}
\caption{Geometric illustration of class margins of various loss functions. (Color for the best view.)}
\label{fig:com_loss}
\end{figure}

In short, GCL in the form of either normalized Euclidean distance or angular distance can achieve the following three advantages: 
1) reduce the softmax saturation and thereby increase the sample validity of tail classes; 
2) avoid overfitting and improve robustness through randomly sampling the values in Gaussian distribution; 
3) enlarge the margin of class boundary for tail classes and thus calibrate the distortion of the embedding space. 
The slight disparity between the two forms lies in the procedural approach: GCL-E incorporates class-based perturbance onto features prior to logit calculation, whereas GCL-A is equivalent to sampling disturbed feature points subsequent to determining their distance from the class anchor.
In addition, we systematically illustrate two versions of GCL and their distinctions from previous methods, exemplified by CE and LDAM~\cite{Kaidi2019ldam}, as depicted in Fig.~\ref{fig:com_loss}.

\subsection{Time-Complexity Analysis}\label{sec:time_ana}
The softmax has a time complexity of $\mathcal{O}(C)$, which is linear with the dimension of logit. It is the same as cross-entropy loss $\mathcal{L}_{CE}$ and $\mathcal{L}_{GCL}$ in both forms. The main difference in time complexity comes from the calculation of logit. For the original normalized logit (which is denoted as $\tilde{z}_j = s \cdot \cos \theta_j$), its main computational cost is vector multiplication. It contains $D\cdot C$ multiplications and $(D-1)\cdot C$ additions. Thus, the time complexity of computing $\tilde{z}_j$ is $\mathcal{O}(DC)$. 
Eq.~(\ref{eq:gcl_ned}) shows that GCL-E only adds $C$ scalar additions to $\tilde{z}_j$. As a result, computing $\tilde{z}^{GCL-E}_j$ has $\mathcal{O}(DC)$ time-complexity. 
For GCL-A, we first expand Eq.~(\ref{eq:gcl_and}) to $\Tilde{z}^{GCL-A}_j=  s\cdot (\cos\theta_j \cos \delta_j\|\varepsilon\| - \sin\theta_j \sin \delta_j\|\varepsilon\| )$. The sine value can be obtained from the corresponding cosine value. Compared to $\tilde{z}_j$, GCL-A adds an additional $2C$ multiplications and $C$ subtractions. Computing $\tilde{z}^{GCL-A}_j$ also has $\mathcal{O}(DC)$ time-complexity. It is obvious that GCL in both forms imposes a negligible additional burden on the training process.

\section{Experiments}\label{sec:exp}
This section first introduces five long-tailed datasets used in our experiments in Section~\ref{sec:dataset}. 
Then, the detailed implementation settings of the experiments are presented in Section~\ref{sec:setting}. 
To demonstrate the effectiveness of GCL, we compare the proposed two forms of GCL with state-of-the-art methods based on a single model structure.
The classification accuracy is compared in Section~\ref{sec:com_res}. 
Moreover, Section~\ref{sec:com_res_moe} validates that GCL can also enhance the performance of MoE model.
Finally, the model validation experiments and ablation studies are conducted to show the properties of our proposed method in Section~\ref{sec:model_val}.

\subsection{Benchmark Datasets}\label{sec:dataset}

We use five benchmarks: CIFAR-10-LT and CIFAR-100-LT, ImageNet-LT, iNaturalist 2018, and Places-LT.

\noindent\textbf{CIFAR-10/100-LT:}
The original versions of CIFAR-10 and CIFAR-100~\cite{krizhevsky2009learning} are uniformly distributed datasets, which consist of 10 and 100 classes, respectively. 
They both contain 60K images with a size of $32\times32$. 
The training set contains 50K samples and the test set has 10K samples. 
Following the experimental settings in~\cite{cui2019class, Kaidi2019ldam}, we down-sampling training images per class with the exponential function $n_i = n^o_{i} \times \lambda^i$, where $i$ is the class index (0-indexed), $n^o_{i}$ is the label frequency in the original balanced version and $\lambda \in (0,1)$. 
The test sets are kept unchanged. The imbalance ratio $r$ is defined as the ratio of the maximum and minimum label frequencies, i.e., $r=\max{(n_i)}/ \min{(n_i)}, i = 1,2,...,C$. 
In the comparative experiments, we employ the three most widely used imbalance ratios, namely $r = 50, 100$, and $200$.

\noindent\textbf{ImageNet-LT and Places-LT:}
The original versions of ImageNet~\cite{ILSVRC15} and Places~\cite{zhou2017places} are artificially balanced, large-scale real-world datasets for classification and localization. 
Following Liu et al.'s \cite{OLTR19}, we construct long-tailed versions of these datasets by truncating a subset using the Pareto distribution with a power value $\alpha=6$ from the balanced versions. 
The original validation sets are employed for testing. 
In summary, ImageNet-LT comprises 115.8K training images from 1K categories with $r=1,280/5$. 
Places-LT consists of 62.5K training images spanning 365 categories with $r=4,980/5$.

\noindent\textbf{iNaturalist 2018:}
iNaturalist 2018 \cite{Horn_2018_CVPR} is a real-world fine-grained dataset for classification and detection, exhibiting a naturally long-tailed distribution.
It contains different species of plants and animals collected from the real world in a wide variety of situations. 
This dataset contains over 437.5K training samples and more than 24.4K validation images from 8,142 categories. 
The official validation set is utilized for testing in the experiments. 
The imbalance ratio of iNaturalist 2018 is $r=1,000/2$.

\subsection{Basic Setting}\label{sec:setting}
The parameters that need to be pre-set are the Gaussian distribution parameters $(\mu, \sigma^2)$. 
For GCL-E, the maximum cloud size cannot exceed 1 because $\cos \theta_i\in [-1,1]$. 
Gaussian distribution has a probability of $99.7\%$ falling in $[\mu-3\sigma, \mu+3\sigma]$, we therefore set $\mu=0$ and $\sigma=\frac{1}{3}$. 
We further clamp $\varepsilon$ to $[-1,1]$ to prevent the cloud size from exceeding 1. For GCL-A, we first constrain the range of $\varepsilon$ to $[-1,1]$ in the same way as the cosine form GCL. 
Then, we multiply $\varepsilon$ with a constant $\frac{\pi}{2}$ to limit the cloud size in angular form to $[-\frac{\pi}{2},\frac{\pi}{2}]$ based on the lemma\footnote{\textbf{Lemma}: The classes can be distributed on a hyper-sphere of dimension $D$ such that any two class centers (namely, class anchors in this paper) are at least $\pi/2$ apart if the number of classes $C$ is less than twice the feature dimension $D$.} proposed by Ranjan et al.~\cite{Ranjan2017L2constrainedSL}. 
Moreover, we normalize $\delta_i$ by $\delta_i \triangleq \delta_i / \max(\delta_{i}), i=1,2,...,C$ to ensure that maximum value of $\delta_i$ does not exceed 1. 
For data augmentation techniques, we follow Zhong et al.~\cite{mislas21}, except for basic augmentation such as image flip, rotation, and random crop, only mixup~\cite{Hongyi2018} are adopted in all experiments to ensure fair comparisons.

PyTorch~\cite{paszke2019pytorch} is utilized to implement the backbone network training. 
We adopt the SGD optimizer with a momentum of 0.9, coupled with a multi-step learning rate schedule.
All models are trained from scratch, except for ResNet-152, which is pre-trained on the original balanced version of ImageNet-1K.
For the first stage, we select ResNet-32 as the backbone network and follow the experimental settings in Cao et al.~\cite{Kaidi2019ldam} for CIFAR-10/100-LT. 
For the experiments conducted on large-scale datasets, namely, ImageNet-LT, iNatralist 2018, and Places-LT, we mainly follow Kang et al.'s settings~\cite{decouple20} except for the learning rate schedule. 
For the second stage, i.e., re-balancing the classifier, we follow Kang~et al.'s setting~\cite{decouple20} for all datasets.

\subsection{Main Comparison Results}\label{sec:com_res}

\begin{table*}[t]
\renewcommand{\thefootnote}{\fnsymbol{footnote}}
 \centering  
 \caption{Comparison results on CIFAR-10/100-LT w.r.t. Top-1 accuracy (\%).} \label{cifar_results}
 \setlength{\tabcolsep}{16pt}
 \renewcommand{\arraystretch}{1.1}
  {\begin{threeparttable}
  \begin{tabular}{l |c c c | c c c}
  \Xhline{0.1em} 
  Dataset & \multicolumn{3}{c|}{CIFAR-10-LT} & \multicolumn{3}{c}{CIFAR-100-LT}\\
  \hline
  \hline 
  Imbalance ratio &200 &100 &50  &200 &100 &50 \\
  \hline 
  CE loss &65.68  &70.70  &74.81  &34.84 &38.43 &43.9\\
  CosFace \cite{wang2018cosface} &66.22 &72.08 & 77.40 & 35.36 & 39.21 & 43.11\\
  ArcFace \cite{Deng2019ArcFace} &66.50 &73.76  &78.19 & 36.64 & 39.06 & 43.40\\
  \cdashline{1-7} 
  LDAM-DRW \cite{Kaidi2019ldam} & 73.52  &77.03  &81.03 &38.91 &42.04 &47.62\\
  De-confound-TDE$^\star$\cite{De-confound-TDE20}    &-  &80.60  &83.60 &- &44.15 &50.31\\
  Decoupling~\cite{decouple20} & 73.06 &79.15 &84.21 & 41.73 & 45.12 & 50.86\\
  BBN~\cite{bbn20}&73.47 &79.82 &81.18 &37.21 &42.56 &47.02\\
  Contrastive learning~\cite{contrastive21}  &- &81.40 & 85.36 &- &46.72 &51.87\\
  MisLAS~\cite{mislas21}   &77.31& 82.06 &85.16 &42.33 &47.50 &52.62 \\
  TSC$^\star$\cite{Li2022TSC} &- &79.70 & 82.90 &- & 43.80 & 47.40\\
  MBJ~\cite{Liu2022Memory}  &77.06 &81.10 & \textbf{85.45} &41.92 & 46.05 & 52.43\\
  FBL~\cite{ICME2022}  &78.10 & 82.46 &84.30 &40.67 &45.22 &50.65 \\
  \cdashline{1-7} 
  {GCL-E (Ours)} &\textbf{79.03} &\underline{\textbf{82.73}} &85.43 &\textbf{44.84} &\textbf{48.69} &\textbf{53.51} \\
  {GCL-A (Ours)} &\underline{\textbf{79.31}} &\textbf{82.72} &\underline{\textbf{85.58}} &\underline{\textbf{46.53}} &\underline{\textbf{49.97}} &\underline{\textbf{54.75}} \\ 
  \Xhline{0.1em}
 \end{tabular}
    \begin{tablenotes}
    \footnotesize
    \item \textbf{Note}: $^\star$ denotes that the results are quoted from the corresponding papers.
    \item Other results are obtained by re-implementing with the official codes.
    \item The best and the second-best results are shown in \underline{\textbf{underline bold}} and \textbf{bold}, respectively.
	\end{tablenotes}
 \end{threeparttable}
 }
\end{table*}
\begin{table}[t]
\renewcommand{\thefootnote}{\fnsymbol{footnote}}
 \centering  
 \caption{Comparison results on ImageNet-LT, iNaturalist 2018 and Places-LT w.r.t. Top-1 accuracy (\%)} \label{large_dataset_results}
 \resizebox{1.\linewidth}{!}
  {\begin{threeparttable}
  \begin{tabular}{l |c|c|c }
  \Xhline{0.1em} &&&\\[-8pt]
  Dataset & img-LT & iNat & Pla-LT\\
  \hline
  \hline &&&\\[-8pt]
  Backbone & ResNet-50 & ResNet-50  & ResNet-152  \\ %
  \hline  &&&\\[-8pt]
  CE loss  &44.51 &63.80   & 27.13 \\
  CosFace \cite{wang2018cosface} &44.95 &72.08 & 27.19 \\
  ArcFace \cite{Deng2019ArcFace}  &44.54 &66.72  & 27.63 \\
  \hdashline  
  LDAM-DRW \cite{Kaidi2019ldam}  &49.96 &68.15 &37.73  \\
  OLTR$^\star$~\cite{OLTR19}  &-  &- &35.90  \\
  Decoupling~\cite{decouple20} &51.68 &70.16 &38.51 \\
  Logit adjustment$^\star$\cite{adjustment21} &51.11 & 66.36 &- \\
  DisAlign$^\star$\cite{DisAli21} &52.91 &70.06 &39.30 \\
  MisLAS \cite{mislas21}    &52.71  &\textbf{71.57} & \textbf{40.36}\\
  TSC$^\star$ \cite{Li2022TSC}   &52.40 & 69.70 & -\\
  MBJ$^\star$ \cite{Liu2022Memory}    &52.10 & 70.00 & 38.10\\
  FBL~\cite{ICME2022}   &50.70  & 69.90 & 38.66\\
  \hdashline &&&\\[-8pt]
  GCL-E (Ours) &\textbf{54.84} &\underline{\textbf{72.01}} &\underline{\textbf{40.62}} \\
  GCL-A (Ours) &\underline{\textbf{55.12}} &71.14 &39.22 \\
  &&&\\[-8pt] \Xhline{0.1em}
 \end{tabular}
   \begin{tablenotes}
    \footnotesize
    \item \textbf{Note}:img-LT, iNat and Pla-LT short for ImageNet-LT, iNaturalist 2018 and Places-LT, respectively. Others are the same as Table~\ref{cifar_results}.
	\end{tablenotes}
   \end{threeparttable}
 }
\end{table}

\subsubsection{Competing Methods}
The competing methods can be categorized into the following two groups.

\noindent\textbf{Baseline Methods:}
Vanilla training with cross-entropy (CE) loss serves as one of our baseline methods. 
Previous studies in visual recognition~\cite{Kaidi2019ldam, Li2018Adaptation, Hou2019VRSTC, Wang2021Deepface} have demonstrated the effectiveness of cosine similarity in mitigating the impact of imbalanced feature bias within imbalanced data distributions. 
Therefore, we also include CosFace~\cite{wang2018cosface} and ArcFace~\cite{Deng2019ArcFace} as additional baseline methods.

\noindent\textbf{State-of-the-art Methods:}
We compare with the most recently proposed state-of-the-art methods, including TSC~\cite{Li2022TSC}, MBJ~\cite{Liu2022Memory}, FBL~\cite{ICME2022}, and two-stage methods including LDAM-DRW~\cite{Kaidi2019ldam} and MisLAS~\cite{mislas21}.
These methods have demonstrated notable classification accuracy across the aforementioned long-tailed datasets. 
For a fair comparison, we implement the experiment of the two-stage strategy, i.e., adding mixup~\cite{Hongyi2018} to decoupling~\cite{decouple20} on all datasets. For CIFAR-10/100-LT datasets, we make a comparison with the logit adjustment method (De-confound-TDE~\cite{De-confound-TDE20}). BBN~\cite{bbn20} and contrastive learning~\cite{contrastive21} are also included in the competing methods. For large-scale datasets, the representation learning method (OLTR~\cite{OLTR19}), and logit adjustment method (logit adjustment~\cite{adjustment21}) are included. The two-stage methods including decoupling~\cite{decouple20}, and DisAlign~\cite{DisAli21} are also compared.

\subsubsection{Comparison Results}
Extensive comparative experiments are conducted to illustrate the efficacy of our proposed GCL in two forms (GCL-E and GCL-A). 
The evaluation metric for assessing performance is top-1 accuracy on the test/validation sets. 
For comparison methods that have not released official code or relevant hyper-parameters, we quote the results directly from the original papers

\noindent\textbf{Results on CIFAR-10/100-LT:}
The proposed GCL-E and GCL-A both outperform the previous methods by notable margins with all imbalanced ratios. 
Especially for the largest $r$, i.e., $200$, the proposed approach has obvious improvement. 
For example, GCL\--E gets $79.03\%$ and $44.84\%$ in top-1 classification accuracy for CIFAR-10-LT and CIFAR\--100\--LT with $r = 200$, which surpasses the second-best method, i.e., FBL~\cite{ICME2022} (on CIFAR-10-LT) and MisLAS~\cite{mislas21} (on CIFAR\--100\--LT) by a significant margin of $0.93\%$ and $2.51\%$, respectively. 
GCL\--A further improves the performance compared to cosine form except on CIFAR-10-LT with $r=100$ ($82.72\%$ top-1 accuracy, which is still higher than the existing methods). For example, it increases the top-1 accuracy from $44.84\%$ to $46.53\%$ for CIFAR-100-LT with $r=200$ compared to the cosine form. 
The margin is more than $3\%$ compared to MisLAS. Interestingly, we can observe that CosFace \cite{wang2018cosface} and ArcFace~\cite{Deng2019ArcFace} perform well compared to CE loss, illustrating the efficacy of angular distance metric in long-tail learning. 
In comparison to LDAM\--DRW~\cite{Kaidi2019ldam} that is also based on angular distance metric, our proposed solution is still the clear winner. 
The performance gain is obtained by the smooth margin that can avoid overfitting and improve robustness. 
The clear performance gain compared to decoupling~\cite{decouple20} demonstrates that calibrating the feature space via GCL is beneficial to the subsequent classifier learning. 
The results on CIFAR-10/100-LT datasets are summarized in Table~\ref{cifar_results}.

\noindent\textbf{Results on Large-scale Datasets:}
The results on large-scale long-tailed datasets including ImageNet-LT, iNaturalist 2018, and Places-LT are reported in Tab.~\ref{large_dataset_results}. 
We observe that GCL-E is superior to the prior arts on all datasets. 
On ImageNet-LT, it achieves $54.84\%$ top-1 accuracy, surpassing DisAlign~\cite{DisAli21} by a notable margin of $1.97\%$ and MisLAS~\cite{mislas21} by 2.77\%. 
For iNaturalist 2018, the proposed GCL-E achieves a top-1 accuracy of 72.01\%, outperforming the second-best method by 0.44\%.
On Place-LT, our proposed method achieves 40.62\% top-1 classification accuracy. 
Although the performance gain compared with MisLAS on iNaturalist 2018 and Place-LT is not as high as other datasets, our method does not require hyper-parameters searching for different datasets and thus is relatively easy to implement. 
GCL-A largely improves the performance on ImageNet-LT from 54.84\% to 55.12\%, but it slightly decreases the accuracy on iNaturalist 2018 and Places-LT. GCL-A achieves 71.14\% top-1 classification accuracy on iNaturalist 2018, which is lower than MisLAS but still outperforms the other baseline methods by notable margins, showing the effectiveness of angular perturbation to balance the embedding space distribution. On Places-LT, it has a lower accuracy than MisLAS and DisAlign.

\subsection{Ablation Study}\label{sec:Abla}

\begin{table}[t]
 \centering  
 \caption{Ablation experiment of different expression forms of cloud size ($\delta^{*}_j$) on CIFAR-10-LT with $r = 100$.} \label{tab:ab_delta}
 \setlength{\tabcolsep}{10pt}
 \resizebox{0.9\linewidth}{!}
  {\begin{tabular}{l|ccc}  
  \Xhline{0.1em} &&&\\[-8pt]
  $\delta^{*}_j$ & Exp. & Expression &Acc(\%)  \\
  \\[-8pt] \hline
  \hline &&&\\[-8pt]
  cos.  & -     &  $\cos (n_j/n_{max} \cdot \pi/2 ) $ & 79.21 \\
  power & $1/3$ & $n_{max} \cdot n^{-1/3}_j$   & 80.80 \\
  power & $1/4$ & $ n_{max} \cdot n^{-1/4}_j$  & 82.31 \\
  log.   & - & $\log n_{max}-\log n_j$  & \underline{\textbf{82.73}}  \\  
  &&&\\[-8pt] \Xhline{0.1em}
 \end{tabular}}
\end{table}

\begin{table}[!t]
 \centering  
 \caption{Ablation experiment of different re-sampling and re-training strategies on CIFAR-10-LT with $r = 100$.} \label{tab:ab_rt_rs}
 \setlength{\tabcolsep}{16pt}
 \resizebox{0.8\linewidth}{!}
  {\begin{tabular}{c c c }
  \Xhline{0.1em} &&\\[-8pt]
  Sampler &RT tech. & Acc.(\%)  \\
  \hline
  \hline &&\\[-8pt]
  IBS   &cRT & 80.52 \\
  SRT   &cRT & 81.74  \\
  ENS   &cRT & 82.45 \\
 \hdashline &&\\[-8pt]
  - & w.o. RT  &80.55  \\
  CBS & LWS   &82.25 \\
  CBS & $\tau$-NC  &82.16 \\
 \hdashline &&\\[-8pt]
  CBS & cRT &\underline{\textbf{82.73}} \\
  &&\\[-8pt] \Xhline{0.1em}
 \end{tabular}}
\end{table}

\noindent\textbf{Expression of Cloud Size:}\label{sec:cloud_size}
We explore several different cloud size adjustment strategies, including power form with different exponents (1/3 and 1/4), and logarithmic form. For a fair comparison, we use GCL-E, and the sampler and re-training strategy are selected as class-balanced sampling and cRT, respectively. The results are presented in Table~\ref{tab:ab_delta}. The logarithmic form has the best performance and the power form with the exponent of 1/4 is also competitive.

\noindent\textbf{Strategies for class re-balancing:} \label{sec:sampler_strategy}
We implement different strategies of data re-sampling and classifier re-training (RT) technique to better analyze our proposed method. 
Table~\ref{tab:ab_rt_rs} shows the results. 
The re-sampling strategy (sampler) includes instance-balanced sampler (IBS)~\cite{decouple20}, square-root sampler (SRS)~\cite{Mahajan2018ECCV}, effective number sampler (ENS)~\cite{cui2019class} and class balanced-sampler (CBS)~\cite{decouple20}. 
The form of GCL is GCL-E and the re-training techniques for all samplers are cRT. 
IBS decreases the performance slightly (from $80.55\%$ to $80.52\%$), which indicates that training the classifier with IBS leads to classifier overfitting. CRT improves the model performance because it increases the sampling probability of tail classes. ENS and CBS have better performance because they can address the problem of negative gradient over suppression by balancing the amount of data in each class. 
We use CBS in the comparison experiments because it achieves the best results among these samplers.
For the selection of RT technique, we first train the backbone without any RT technology using GCL-E. 
Then we froze the representation and re-balance the classifier with learnable weight scaling (LWS), $\tau$-normalized classifier ($\tau$-NC), and cRT, respectively. 
We can observe that even without any RT technique, our approach (the top-1 classification accuracy is $80.55\%$) can still beat most state-of-the-art including two-stage methods (for example, LDAM-DRW and BBN achieve $77.03\%$ and $79.82\%$, respectively). All RT techniques significantly improve model performance, which demonstrates that good representation can improve classification accuracy by simply re-balancing the classifier. cRT outperforms best among the classifier re-training techniques, which improves the accuracy by $2.18\%$ compared with no RT. Thus, we use cRT in the comparison experiments.

\subsection{Further Analysis}\label{sec:model_val}

We conduct a series of experiments to further analyze the proposed method.

\noindent\textbf{Effectiveness on MoE model:}\label{sec:com_res_moe}
We select RIDE~\cite{Wang21ride} as a representative of MoE Models.
The reproduction of RIDE in our experiment follows the original settings, which utilize LDAM loss and DRW strategy. 
We employed three experts in our MoE model and adopted the mixup technique to ensure a fair comparison.
MoE models have been shown to outperform single models, albeit at the expense of increasing model size. 
For instance, RIDE with GCL-E achieved an accuracy of 81.32$\%$ on CIFAR-10-LT with an imbalance ratio of 200, which is an obvious improvement from the 79.03$\%$ achieved by a single ResNet-32 model with GCL-E. 
However, the model size of RIDE is 5.38 Mb, whereas the single model had a size of only 1.84 Mb.
Tables~\ref{cifar_ride} and \ref{large_dataset_ride} demonstrate the improvement in performance achieved by GCL on RIDE.
Both versions of GCL can be observed to improve RIDE's performance significantly on all datasets.
The improvement of GCL-A ranges from 0.90$\%$ to 2.62$\%$, while that of GCL-E ranges from 0.82$\%$ to 2.64$\%$.

\begin{table*}[t]
\renewcommand{\thefootnote}{\fnsymbol{footnote}}
 \centering  
 \caption{Validation of the effect on MoE model on CIFAR-10/100-LT.} \label{cifar_ride}
 \resizebox{0.95\textwidth}{!}
 {
  \begin{tabular}{l |c c c | c c c}
  \Xhline{0.1em} &&&&&&\\[-8pt]
  Dataset & \multicolumn{3}{|c|}{CIFAR-10-LT} & \multicolumn{3}{|c}{CIFAR-100-LT}\\
  \hline
  \hline \\[-8pt]
  Backbone & \multicolumn{6}{c}{ResNet-32}\\ %
  \hline &&&&&&\\[-8pt]
  Imbalance ratio &200 &100 &50  &200 &100 &50 \\
  \hline &&&&&&\\[-8pt]
  RIDE \cite{Wang21ride} & 80.42 & 83.39 & 85.34 &  47.80 & 50.91 & 54.87\\
  \hdashline &&&&&&\\[-8pt]
  {RIDE w. GCL-E} & 81.32 \color{blue}($\uparrow $ 0.90) & 84.32 \color{blue}($\uparrow $ 0.93) & 87.03 \color{blue}($\uparrow$ 1.69) & 48.96 \color{blue}($\uparrow $ 1.16) & 52.57 \color{blue}($\uparrow $ 1.66) &  57.49 \color{blue}($\uparrow $ 2.62) \\
  {RIDE w. GCL-A} &82.08 \color{blue}($\uparrow $ 1.66) & 84.73 \color{blue}($\uparrow $ 1.34) & 86.95 \color{blue}($\uparrow $ 1.61) & 48.62 \color{blue}($\uparrow $ 0.82) & 52.38 \color{blue}($\uparrow $ 1.47) &  57.51 \color{blue}($\uparrow $ 2.64) \\ 
  &&&&&&\\[-8pt] \Xhline{0.1em}
 \end{tabular} 
 }
\end{table*}
\begin{table}[t]
\renewcommand{\thefootnote}{\fnsymbol{footnote}}
 \centering  
 \caption{Validation of the effect on MoE model on large-scale dataset.} \label{large_dataset_ride}
 \resizebox{1.\linewidth}{!}
  {
  \begin{tabular}{l |c|c|c }
  \Xhline{0.1em} &&&\\[-8pt]
  Dataset & ImageNet-LT & iNaturalist 2018 & Places-LT\\
  \hline
  \hline &&&\\[-8pt]
  Backbone & ResNet-50 & ResNet-50  & ResNet-152  \\ %
  \hline  &&&\\[-8pt]
   RIDE \cite{Wang21ride}  &55.55 &72.17 & 39.91 \\
  \hdashline  &&&\\[-8pt]
  {RIDE w. GCL-E}  &57.01 \color{blue}($\uparrow $ 1.46) &74.27 \color{blue}($\uparrow $ 2.10) & 41.06 \color{blue} ($\uparrow $ 1.15)  \\
 {RIDE w. GCL-A}  & 57.25 \color{blue}($\uparrow $ 1.70) & 73.56 \color{blue}($\uparrow $ 1.39) & 41.50 \color{blue} ($\uparrow $ 1.59) \\
  &&&\\[-8pt] \Xhline{0.1em}
\end{tabular}
}
\end{table}

\begin{figure}[tb]
  \centering
  \subfloat[$\delta=0.1$]{
  \hspace{-10pt}
        \includegraphics[width=0.5\linewidth]{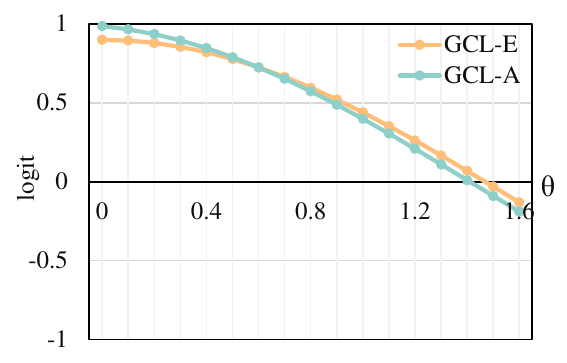}
        \label{fig:curve01}}
  \subfloat[$\delta=0.5$]{
        \includegraphics[width=0.5\linewidth]{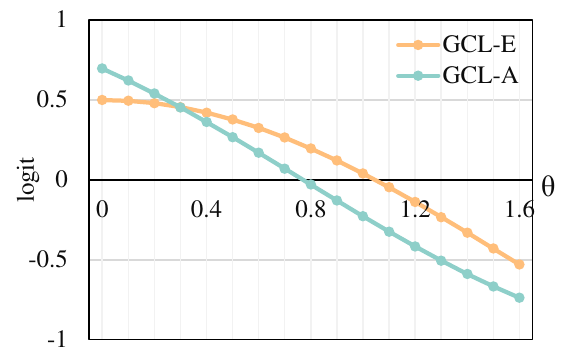}
        \label{fig:curve05}}
\caption{The logit curve of GCL in different forms. For ease of visualization, the scale parameter $s$ is omitted\protect\footnotemark.}
\label{fig:NEDvsAND}
\vspace{-6pt}
\end{figure}

\noindent\textbf{GCL-E vs. GCL-A:}
Combining Tables~\ref{cifar_results} and \ref{large_dataset_results}, it can be observed that GCL-A does not always have inferior performance compared to GCL-E, and vice versa.
The reason is that iNaturalist 2018 and Places-LT have much large imbalance ratios ($r=500$ and $996$, respectively) than the other datasets (ImageNet-LT has the largest $r$ which is 256 among these datasets ). 
We draw the logit curve of different forms of GCL, which is shown in Fig.~\ref{fig:NEDvsAND}. 
In our setting, the large class has a small $\delta$. 
The smaller the class size, the larger its corresponding $\delta$. 
As the distance $\theta$ increases, the logit of GCL-A decreases faster than GCL-E. 
It is more noticeable for the larger $\delta$, as shown in Fig.~\ref{fig:curve05}. 
A small distance will have a more obvious logit difference for GCL-A compared with GCL-E. 
Therefore, in the case of high imbalance ratio, GCL-E can make the separability of minority classes stronger so that the logit difference is more significant.

Another rationale arises from the discrepancy in logits restrictions caused by varying imbalance ratios.
Excessively strict logit constraints may lead the model astray. 
Without loss of generality, we use the most frequent class (denoted by subscript `head') and the least frequent class (denoted by subscript `tail') to analyze.
For an input image that is tail class, GCL-A necessitates:
\begin{equation}
\begin{array}{cc}
    \cos (\theta_{tail}+ \delta \cdot \dfrac{\pi}{2} \cdot \|\varepsilon\|)  > \cos \theta_{head} \Rightarrow \\ 
     \theta_{tail}   < \theta_{head} - \delta \cdot \dfrac{\pi}{2} \cdot \|\varepsilon\|.
\end{array}
\end{equation}
Considering $\delta=0.5$ as an example, when $\theta_{head} < \frac{\pi}{2}$, $\theta_{tail}$ being negative satisfies the requirements of the loss function, which could mislead the model training.
The requirement that the angle between non-target classes and the target weight be greater than $\frac{\pi}{2}$ is overly stringent.
For highly imbalanced datasets, namely iNaturalist 2018 and Places-LT, the discrepancies in perturbations between tail and head classes are more pronounced, which contributes to this phenomenon. 
In datasets with a smaller imbalance ratio, the disparities in perturbations are comparatively smaller, making this restriction relatively weaker.
The majority of classes can adhere to their respective soft margin restrictions.
However, opting for a smaller $\delta$ might result in the added perturbation being less conspicuous, thereby leading to less differentiation between classes.
For GCL-E, an input image belonging to the tail class should satisfy the following inequality:
\begin{equation}
\begin{array}{cc}
    \cos \theta_{tail} - \delta \cdot \|\varepsilon\|  > \cos \theta_{head} \Rightarrow \\ 
    \cos \theta_{tail}  < \cos \theta_{head} + \delta \cdot \|\varepsilon\|.
\end{array}
\end{equation}
When $\delta=0.5$, $\theta_{head} > \frac{\pi}{3}$ will cause $\theta_{tail}$ to be negative. 
In contrast, the constraints imposed by GCL-E are more lenient, resulting in a slight decrease in performance on datasets characterized by a low imbalance ratio compared to GCL-A.
Nonetheless, this relaxation does not predispose the model to erroneous interpretations stemming from excessively stringent restrictions.

Moreover, from another perspective, the selection of the perturbation magnitude $\delta$ holds a pivotal role for GCL-A. 
Additionally, cloud size selection should extend beyond mere class size considerations, with each variant of GCL potentially requiring its optimal strategy for cloud size selection.
It is conceivable that the logarithmic form of cloud size utilized for GCL-A does not constitute the optimal choice. 
We leave these as our future study.

\footnotetext{Specifically, the logit curves show $\Tilde{z}^{GCL-E}=\cos(\theta )-\delta\cdot\|\varepsilon\|, \text{ and } \Tilde{z}^{GCL-A}= \cos(\theta + \delta \cdot \frac{\pi}{2} \cdot  \|\varepsilon\|)$, namely $s=1$.}

\begin{figure*}[ht]
  \centering
  \subfloat[CE loss]{
  \hspace{-12pt}
        \includegraphics[width=0.48\textwidth]{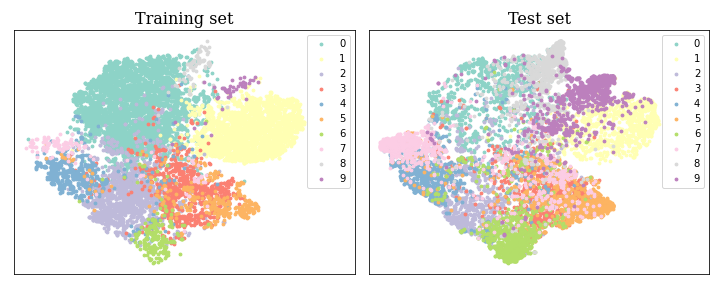}
        \label{fig:tsne-ce}}
  \subfloat[LDAM loss]{
        \includegraphics[width=0.48\textwidth]{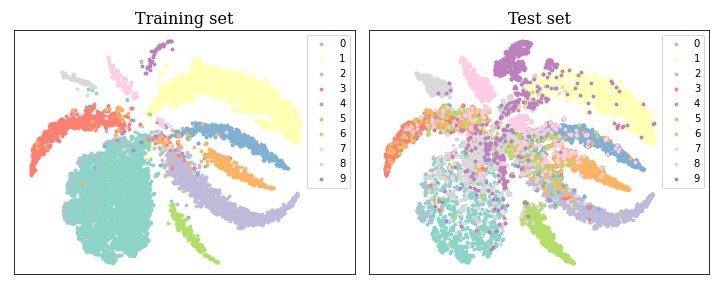}
        \label{fig:tsne-ldam}}
  \\
  \subfloat[GCL-E]{
  \hspace{-12pt}
        \includegraphics[width=0.48\textwidth]{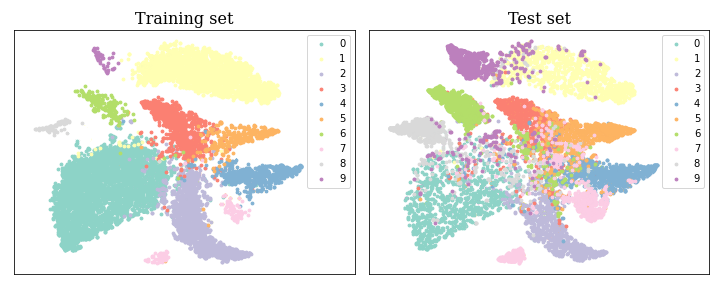}
        \label{fig:tsne-gclc}}
  \subfloat[GCL-A]{
        \includegraphics[width=0.48\textwidth]{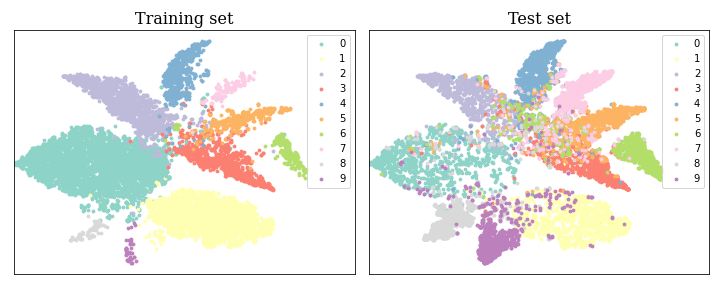}
        \label{fig:tsne-gcla}}
\caption{Visualization of the embedding distribution obtained by different methods. t-SNE projection is utilized. The dataset is CIFAR-10-LT with $r = 100$. ResNet-32 is used as the backbone. (Color for the best view.)}
\label{fig:visualization}
\end{figure*}

\begin{figure}[ht]
  \centering
  \subfloat[CE loss]{
        \hspace{-12pt}
        \includegraphics[width=0.24\textwidth]{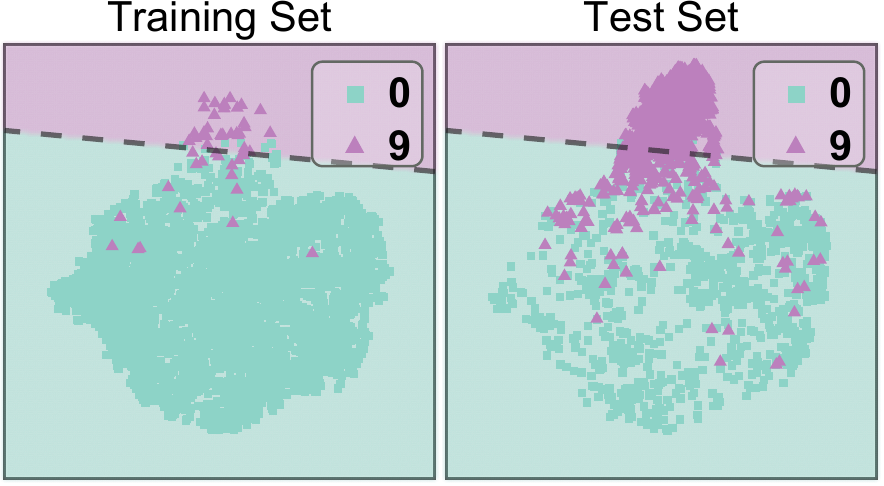}
        \label{fig:tsne-ce09}}
  \subfloat[LDAM loss]{
        \includegraphics[width=0.24\textwidth]{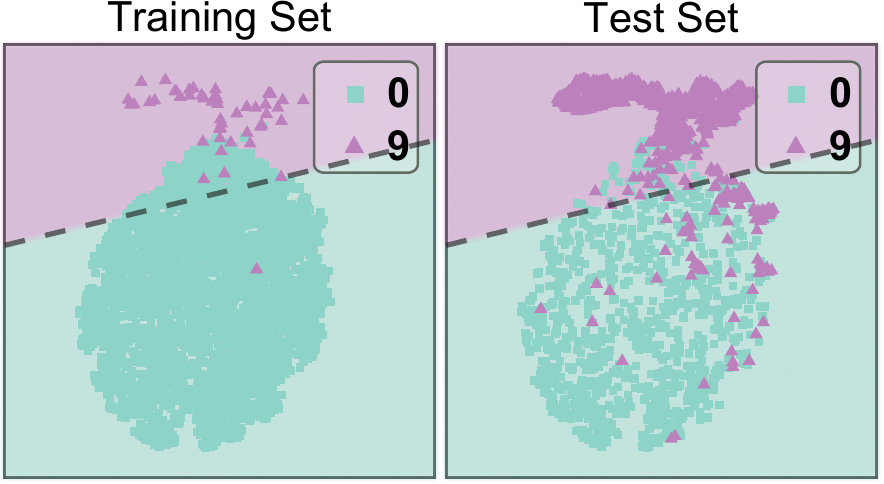}
        \label{fig:tsne-ldam09}}
  \\[-6pt]
  \subfloat[GCL-E]{
         \hspace{-12pt}
         \includegraphics[width=0.24\textwidth]{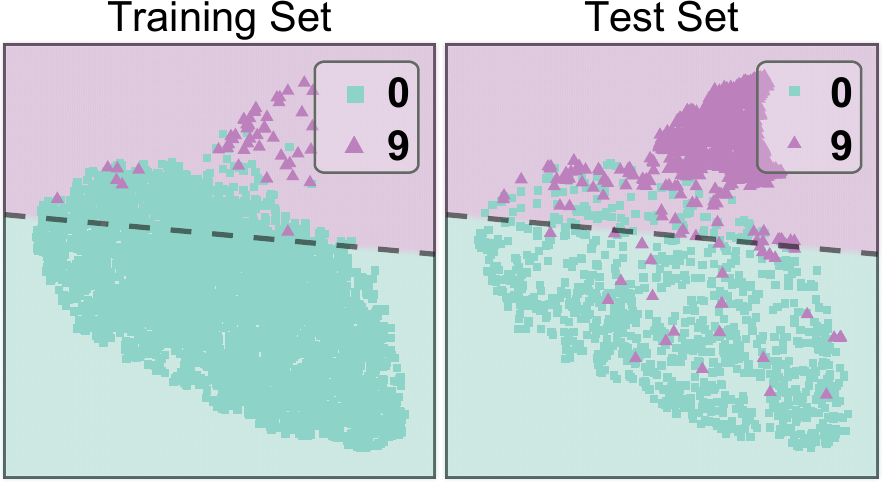}
        \label{fig:tsne-gclc09}}
  \subfloat[GCL-A]{
        \includegraphics[width=0.24\textwidth]{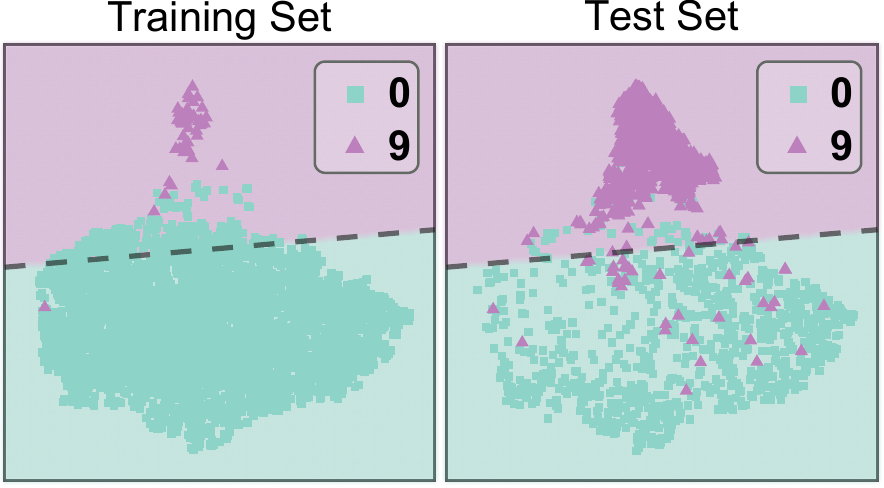}
        \label{fig:tsne-gcla09}}
\caption{T-SNE visualization of decision boundary (dashed line) between head (class 0) and tail (class 9) classes. 
The dataset is CIFAR-10-LT with $r = 100$ and the backbone network is ResNet-32. (Color for the best view.)}
\label{fig:vis-2class}
\end{figure}

\begin{table}[tb]
    \centering
    \caption{Top-1 Classification Accuracy ($\%$) of the three splits on ImageNet-LT.}
    \setlength{\tabcolsep}{6pt}
    \resizebox{\linewidth}{!}
    {
    \begin{tabular}{l|cccc}
    \Xhline{0.1em} &&&&\\[-8pt]
    Method & Head & Middle & Tail & Overall\\
    \hline
    \hline &&&&\\[-8pt]
    Class size & $n_j>100 $ & $20 < n_j \leq 100$ & $n_j \leq20$ & -\\
    \hline &&&&\\[-8pt]
    CE    & \underline{\textbf{64.91}} & 38.10  & 11.28 & 44.51 \\
    CosFace & 64.48 & 39.26  & 11.55 & 44.95\\
    ArcFace & \textbf{64.86} & 38.07  & 11.75 & 44.54\\
    \hdashline &&&&\\[-8pt]
    LDAM-DRW & 58.63 & 48.95  & 30.37 & 49.96\\
    OLTR & 61.93 & 44.68  & 19.98 & 47.72\\
    Decoupling  & 63.71 & 43.01  & 20.55 & 47.70\\
    MisLAS & 62.43  &49.31 & 33.89 & 52.11\\
    \hdashline &&&&\\[-8pt]
    GCL-E & 63.78 &\textbf{52.62} & \textbf{38.70} & \textbf{54.84}\\
    GCL-A & 62.72 &\underline{\textbf{53.26}} & \underline{\textbf{40.95}} & \underline{\textbf{55.12}}\\
    &&&&\\[-8pt] \Xhline{0.1em}
    \end{tabular}
    }\label{tab:acc_spl}
\end{table}

\noindent\textbf{The Effect of Gaussian Cloud:}
To obtain additional insight, we visualize the embedding distribution using t-SNE projection. Since CE loss is selected as the loss function for several methods~\cite{bbn20, decouple20, OLTR19}, especially MisLAS performs the second-best in most cases, we visualize the embedding distribution obtained by CE loss for comparison. LDAM~\cite{Kaidi2019ldam} is an angular distance metric based method but utilizes the hard margin, we also show its embedding distribution. The embeddings are calculated from the samples in CIFAR-10-LT with $r = 100$. Fig.~\ref{fig:visualization} shows the results. From Fig.~\ref{fig:tsne-ce}, it can be seen that the embeddings of each class obtained via CE loss are clustered together and are relatively difficult to separate. The obscure region of CE loss embedding is larger than that of other approaches. LDAM and GCL in both forms are all angular distance metric based methods, thus their embeddings are basically radial. Fig.~\ref{fig:tsne-ldam} shows that the LDAM embedding of each class is more slender. This is caused by the hard margin that strictly restricts the class region, resulting in overfitting the training set. Thus, LDAM does not generalize well on the test set compared with our proposed GCL. In Fig.~\ref{fig:tsne-gclc} and Fig.~\ref{fig:tsne-gcla}, on training set, the embeddings for each class obtained via GCL in both forms have more obvious margins compared to CE and also are more scattered compared to LDAM. The results of the test set verify the efficacy of our proposed approach. GCL-E and GCL-A have better generalization performance, and it can be found that the misclassified classes are mainly in the edge regions of each class. 
For better illustration, we additionally compare the embedding distribution of the most (class 0) and least (class 9) frequent classes, along with their respective decision boundaries derived from various loss functions in Fig.~\ref{fig:vis-2class}. 
Concerning the acquired features, within the training set, the overlap between the features of the head and tail classes by LDAM and GCL is reduced compared to those obtained by CE loss, with a pronounced disparity observed in GCL-A.
In addition, it presents more clearly that compared to our proposed GCL, the LDAM embeddings appear to perform better on the training set, but cannot be well generalized to the unseen test samples. 
In Fig.~\ref{fig:tsne-ldam09}, there are more points of class 9 appearing inside the class 0 area on the test set. 
By contrast, as shown in Fig.~\ref{fig:tsne-gclc09} and Fig.~\ref{fig:tsne-gcla09}, the misclassified points of class 9 are mainly in the edge area of class 0 on test set.
Regarding the decision boundary, CE loss exhibits a tendency to predominantly ensure accurate classification of head classes while often disregarding tail classes.
In contrast, due to the presence of margins or perturbations beneficial to the tail class, both LDAM and GCL adopt a holistic approach to class performance.
However, this approach comes at the expense of head class performance to some extent.
The decision boundary delineates specific head class samples into the tail class.

\noindent\textbf{Performance on Classes with Different Scale:} \label{sec:group_performance}
To investigate the impact of GCL, we report the accuracy of various scale classes on ImageNet-LT. 
The results are presented in Table~\ref{tab:acc_spl}. 
The classification accuracy of baseline methods drops a lot in the middle and tail classes. LDAM-DRW increases the accuracy of middle and tail classes but decreases that of head classes a lot. 
GCL-E outperforms the other state-of-the-art methods on middle and tail classes with large margins. 
Meanwhile, the accuracy of the head class decreases the least. By contrast, GCL-A has more improvement in middle and tail classes, but the damage to head classes is slightly higher than GCL-E and decoupling. In general, GCL-E performs well in all class scales. GCL-A has the highest overall classification accuracy. Significantly improving the accuracy of tail classes while preventing that of the head classes from diminishing illustrates the superiority of our approach.

\section{Conclusion}\label{sec:conclusion}
In this paper, we have proposed to use Gaussian form perturbance to augment the features for long-tailed classification.
Eventually, we have derived two GCL forms, which are simple but effective.
Both of these two forms make tail classes have larger perturbance amplitudes on their corresponding class anchors, which can expand the spatial distribution of tail class embeddings.
Furthermore, we have analyzed the rationale of the proposed method from different perspectives, which provides insights into how to obtain a representative and balanced-distributed embedding.
After obtaining a balanced distributed embedding space, the classifier bias can be effectively addressed by simply retraining it with class-balanced sampling.
Comprehensive experiments on various benchmark datasets have demonstrated that the proposed Gaussian clouded logit in both forms achieves significant performance gains compared to the state-of-the-art methods.
In addition, we have also validated the properties of the proposed GCL by t-SNE visualization and the performance on different scales of classes.


\bibliographystyle{IEEEtran}
\bibliography{IEEEfull,reference}


\begin{IEEEbiography}[{\includegraphics[width=1in,height=1.25in,clip,keepaspectratio]{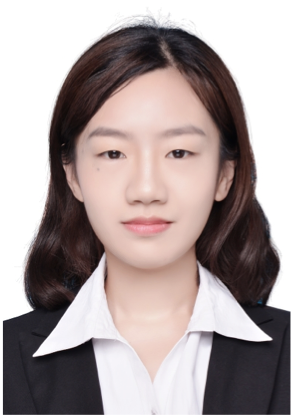}}]{Mengke Li}
received the B.S. degree in Communication Engineering from Southwest University, Chongqing, China, in 2015, the M.S. degree in electronic engineering from Xidian University, Xi’an, China, in 2018, and the Ph.D. degree from the Department of Computer Science, Hong Kong Baptist University, Hong Kong SAR, China, under the supervision of Prof. Yiu-ming Cheung, in 2022. She is currently an Associate Researcher with Guangdong Laboratory of Artificial Intelligence and Digital Economy (SZ), Guangdong, China. Her current research interests include imbalanced data learning, long-tail learning and pattern recognition.
\end{IEEEbiography}

\begin{IEEEbiography}[{\includegraphics[width=1in,height=1.25in,clip,keepaspectratio]{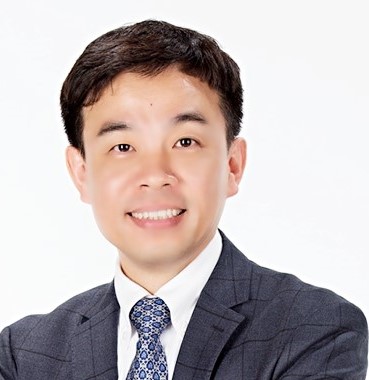}}]{Yiu-ming Cheung}
(SM'06-F'18) received the Ph.D. degree from the Department of Computer Science and Engineering at The Chinese University of Hong Kong in Hong Kong. He is a Fellow of IEEE, AAAS, IET, BCS, and AAIA. He is currently a Chair Professor (Artificial Intelligence) of the Department of Computer Science, Hong Kong Baptist University, Hong Kong SAR, China. His research interests include machine learning, visual computing, data science, pattern recognition, multi-objective optimization, and information security. He is currently the Editor-in-Chief of IEEE Transactions on Emerging Topics in Computational Intelligence. Also, he serves as an Associate Editor for IEEE Transactions on Cybernetics, IEEE Transactions on Cognitive and Developmental Systems, IEEE Transactions on Neural Networks and Learning Systems (2014-2020), Pattern Recognition, Pattern Recognition Letters, and Neurocomputing, to name a few. For details, please refer to: \url{https://www.comp.hkbu.edu.hk/~ymc}.
\end{IEEEbiography}

\begin{IEEEbiography}[{\includegraphics[width=1in,height=1.25in,clip,keepaspectratio]{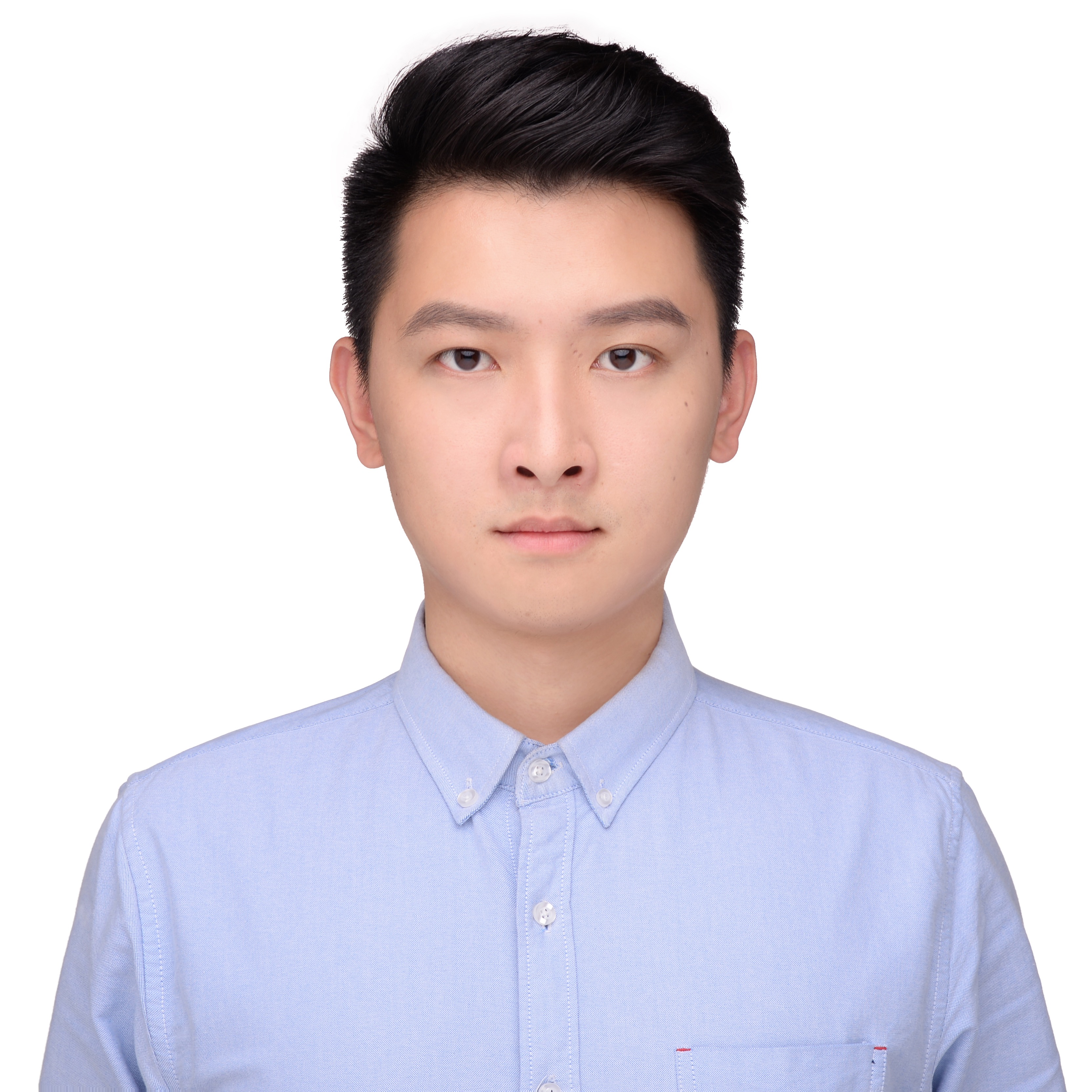}}]{Yang Lu}
received the B.Sc. and M.Sc. degrees in Software Engineering from the University of Macau, Macau, China, in 2012 and 2014, respectively, and the Ph.D. degree in computer science from Hong Kong Baptist University, Hong Kong, China, in 2019. He is currently an Assistant Professor with the Department of Computer Science, School of Informatics, Xiamen University, Xiamen, China. His current research interests include deep learning, federated learning, long-tail learning, and meta-learning.
\end{IEEEbiography}

\begin{IEEEbiography}[{\includegraphics[width=1in,height=1.25in,clip,keepaspectratio]{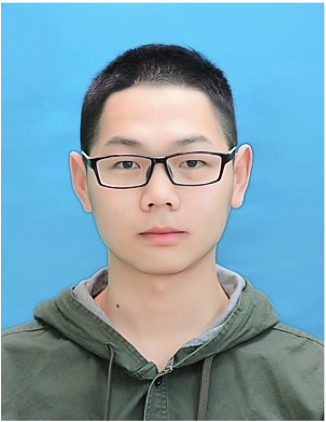}}]{Zhikai Hu}
received his B.S. degree in Computer Science from China Jiliang University, Hangzhou, China, in 2015, and the M.S. degree in computer science from  Huaqiao University, Xiamen, China, in 2019. He is currently pursuing the Ph.D. degree with the Department of Computer Science, Hong Kong Baptist University, Hong Kong SAR, China, under the supervision of Prof. Yiu-ming Cheung. His present research interests include information retrieval, pattern recognition and data mining.
\end{IEEEbiography}

\begin{IEEEbiography}[{\includegraphics[width=1in,height=1.25in,clip,keepaspectratio]{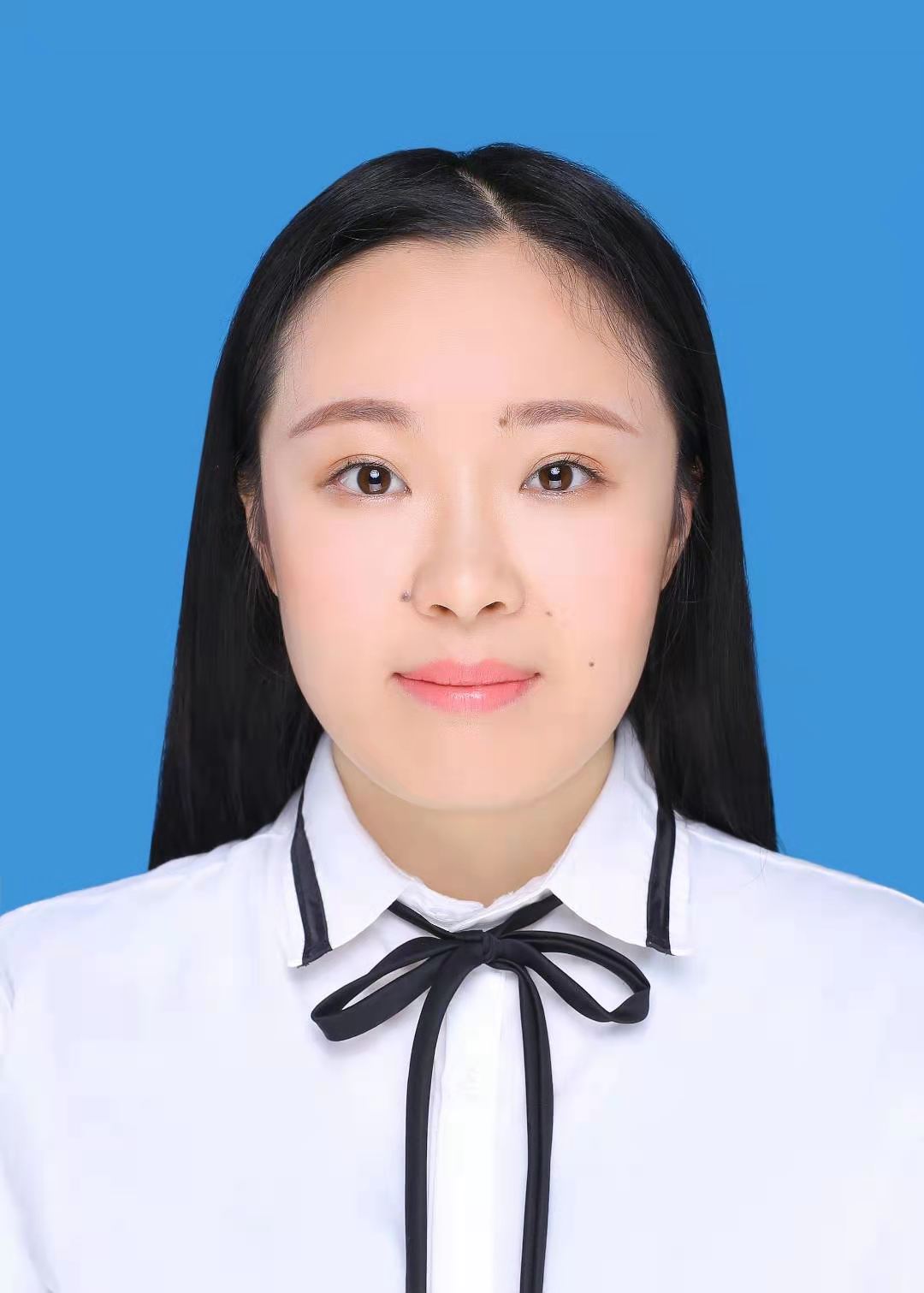}}]{Weichao Lan}
received her B.S. degree in Electronics and Information Engineering from Sichuan University, Chengdu, China, in 2019, and the Ph.D degree in computer science from Hong Kong Baptist University, Hong Kong SAR, China, under the supervision of Prof. Yiu-ming Cheung, in 2024. Her present research interests include network compression and acceleration, lightweight models.
\end{IEEEbiography}

\begin{IEEEbiography}[{\includegraphics[width=1in,height=1.25in,clip,keepaspectratio]{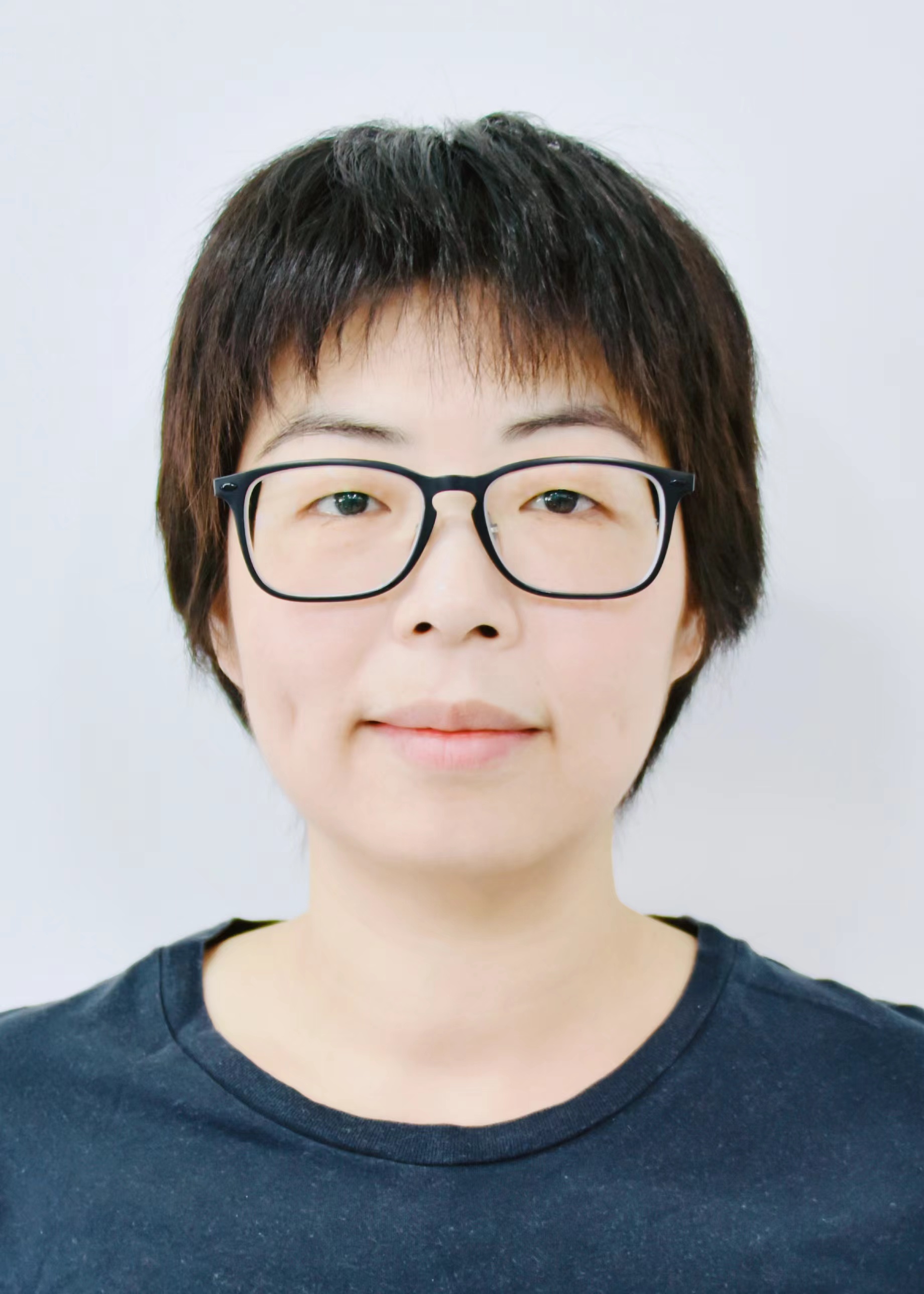}}]{Hui Huang}
(Senior Member, IEEE) received the Ph.D. degree in applied mathematics from The University of British Columbia in 2008. She is a Distinguished Professor at Shenzhen University, serving as the Dean of College of Computer Science and Software Engineering while also directing the Visual Computing Research Center. Her research encompasses computer graphics, computer vision and visual analytics, focusing on geometry, points, shapes and images. She is currently on the editorial board of ACM TOG and IEEE TVCG.

\end{IEEEbiography}

\end{document}